\pdfoutput=1

\documentclass[11pt]{article}

\usepackage[final]{acl}

\usepackage{times}
\usepackage{latexsym}

\usepackage[T1]{fontenc}

\usepackage[utf8]{inputenc}

\usepackage{microtype}

\usepackage{inconsolata}

\usepackage{graphicx}
\usepackage{float}
\usepackage{enumitem}
\usepackage{verbatim}
\usepackage{hyperref}
\usepackage{blindtext}
\usepackage{booktabs}
\usepackage{multirow}
\usepackage{array}
\usepackage{subcaption}
\usepackage{amsmath}
\usepackage{amssymb}
\usepackage{paralist}
\usepackage{vcell}
\usepackage{xcolor,lipsum,subcaption}
\usepackage{tabularx}
\usepackage{rotating, adjustbox}

\usepackage{pifont}

\newcommand{\cmark}{\textcolor{cyan}{\ding{51}}}
\newcommand{\xmark}{\textcolor{red}{\ding{55}}}

\title{A Multi-Labeled Dataset for Indonesian Discourse: \\ Examining Toxicity, Polarization, and Demographics Information}
\author{
    \textnormal{Lucky Susanto$^{*,1}$ \quad Musa Wijanarko$^{*,1}$ \quad Prasetia Pratama$^{4}$ \quad Zilu Tang$^{2}$} \\
    \textnormal{Fariz Akyas$^{1}$ \quad Traci Hong$^{2}$ \quad Ika Idris$^{1}$ \quad Alham Aji$^{3}$ \quad Derry Wijaya$^{1}$}\\
    \textnormal{$^1$Monash University Indonesia \quad $^2$Boston University \quad $^3$MBZUAI}\\
    \textnormal{$^4$Independent Researcher\quad $^*$Equal Contributor}\\
}

\begin{document}
\maketitle
\begin{abstract}
Polarization is defined as divisive opinions held by two or more groups on substantive issues. As the world’s third-largest democracy, Indonesia faces growing concerns about the interplay between political polarization and online toxicity, which is often directed at vulnerable minority groups. Despite the importance of this issue, previous NLP research has not fully explored the relationship between toxicity and polarization. To bridge this gap, we present a novel multi-label Indonesian dataset that incorporates toxicity, polarization, and annotator demographic information. Benchmarking this dataset using BERT-base models and large language models (LLMs) shows that polarization information enhances toxicity classification, and vice versa. Furthermore, providing demographic information significantly improves the performance of polarization classification.

\end{abstract}


\section{Introduction}

Political polarization and online toxicity are growing global concerns, especially during politically charged events. While ideological differences are natural in a healthy democracy, extreme polarization deepens divisions, often escalating into hostility and societal fragmentation \citep{mccoy2018pernicious}. In such cases, opposing groups begin to see each other as existential threats, making reconciliation increasingly difficult \citep{kolod2024polarization, milavcic2021negative}. At the same time, online toxicity disproportionately affects minority groups \citep{alexandra-2023}, leading to self-censorship \citep{toxicity-self-censorship} and undermining public discourse, particularly in journalism \citep{toxicity-threatens-journalism, hatespeech-williams-2019}.

Indonesia, the world's third-largest democracy, is home to 277 million people from diverse backgrounds \citep{datacommons-indonesia-population}, making it a significant case study. The 2024 presidential election saw intense political competition alongside a sharp rise in divisive and toxic online discussions. For example, \citet{csis} found that in 2019, 1.35\% of 800,000 online texts contained toxic language, while by 2024, \citet{aji-hatespeech-dashboard} reported that 13.8\% of 1.45 million texts were toxic—a tenfold increase. This surge highlights the growing toxicity in Indonesian discourse. However, despite the high-stakes nature of Indonesian elections, the relationship between political polarization and online toxicity has not been thoroughly examined.

Extensive research has explored toxicity and polarization separately, yet their complex relationship remains largely unstudied. This gap limits our understanding of how hostile online environments evolve. While political polarization can intensify toxicity, not all polarized discourse is toxic, and not all toxic speech is politically polarized. A dataset that captures both distinctions allows for a clearer differentiation between divisive yet civil discussions and interactions that escalate into outright hostility. To address this, \textbf{we introduce the first multi-labeled Indonesian dataset that includes toxicity, polarization, and annotator demographic information}\footnote{Dataset available at \url{https://huggingface.co/datasets/Exqrch/IndoToxic2024}}. This dataset serves as a foundation for investigating how these factors interact in online discourse, offering insights into the broader implications of digital polarization and toxicity.


\section{Interplay Of Toxicity, Political Polarization, and Identity}
Online discourse is increasingly characterized by a vicious cycle in which polarization fuels toxic language and vice versa. Social media platforms exacerbate these dynamics by enabling unopposed expression of opinions, thereby deepening societal divisions \citep{PaG6663, Vasist2024, schweighofer2018polarization}. 

\subsection{Toxicity and Polarization}
Online discussions increasingly follow a vicious cycle in which polarization fuels toxic language, and toxicity, in turn, reinforces polarization. Social media platforms intensify these dynamics by allowing unchallenged expression of opinions, reinforcing existing beliefs and deepening societal divides \citep{PaG6663, Vasist2024, schweighofer2018polarization}. These effects are particularly pronounced in highly polarized political environments, where online interactions often escalate from ideological disagreements to outright hostility.

Specifically for polarization, recent work has shifted focus from ideological to identity-based polarization \citep{schweighofer2018polarization}. While political polarization is defined as a divide in the population between political groups on either side of the political orientation spectrum \cite{weber2021-political-polarization}. Polarizing messages, driven to reinforce inter-group biases and invoke a strong in-group identity, occasionally take the form of toxicity, as defined by \citet{Donohue2022}. While the converse is also true (see Appendix \ref{app:example-5sample}), the two phenomena remain distinct.

\subsection{Non-Toxic Polarization}
Diverse opinions are essential to democracy \citep{powell2022toxic}. Yet, without a willingness to compromise \citep{compromise}, even civil exchanges can generate polarization. This non-toxic polarization may erode common ground \citep{dimaggio1996polarized}, foster echo chambers \citep{echo-chamber}, and normalize extreme positions \citep{extremism}.

\subsection{How Identities Shape Discourse Dynamics}
Identity plays a pivotal role in shaping online discourse by influencing both opinion formation and interaction patterns. Research shows that identity issues are among the strongest drivers of polarization \citep{milavcic2021negative}. In diverse societies, variations in cultural, social, and political identities can intensify divisions. Initially, exposure to diversity may reduce both in-group and out-group trust \citep{putnam-2007}, undermining constructive dialogue. Moreover, heightened polarization is often linked with increased online toxicity, frequently directed at vulnerable and minority groups \citep{alexandra-2023}. However, \citet{putnam-2007} also state that sustained outer-group interaction beyond a critical threshold can foster inclusive encompassing identities and potentially mitigate polarization.

In summary, the interplay between toxicity, polarization, and demographic identities remains a critical yet understudied aspect of online discourse. By integrating demographic factors into our analysis, we aim to provide a nuanced understanding of how identities shape discourse dynamics and develop targeted strategies for mitigating both polarization and toxicity in digital environments.

\section{Available Datasets}
\paragraph{Polarization Datasets} Most polarization datasets have been developed from U.S.-centric studies \cite{khudabukhsh2021we, sinno2022political}. However, recent work has expanded this focus to include non-U.S. contexts. For instance, \citet{vorakitphan-etal-2020-regrexit} introduced a dataset examining polarization during the Brexit phenomenon by analyzing partisan news media in England. In addition, \citet{szwoch-etal-2022-creation} compiled a dataset on polarization in Poland by analyzing articles from both state-owned and commercial media.

\paragraph{Toxicity Datasets}
A variety of datasets have been developed to detect and analyze online toxicity. For example, \citet{kumar-2021} employ a continuous scale to measure toxicity, whereas \citet{Davidson_Warmsley_Macy_Weber_2017} introduced a dataset categorizing content as \textit{Hate}, \textit{Offensive}, or \textit{Neither}. More recently, toxicity datasets for relatively low-resource languages have emerged, such as Brazilian Portuguese \citep{lima-etal-2024-toxic}; Vietnamese \citep{hoang-etal-2023-vihos}; and Korean \citep{moon-etal-2020-beep}, which are crucial for advancing automatic moderation tools.

\paragraph{Toxicity and Polarization Dataset}
While prior work has examined polarization and toxicity separately, our dataset is the first to provide multi-label annotations for both, enabling nuanced analysis of their intersection in a non-Western context. A full comparison of available datasets is provided in Appendix \ref{app:dataset-comparison}. 

\section{Dataset Creation}

\begin{table}[ht]
\small
\centering
\resizebox{\columnwidth}{!}{
\begin{tabular}[ht]{lrr}
\toprule
\textbf{Demographic} & \textbf{Group}& \textbf{Count}\\
\midrule
Disability  & \begin{tabular}[t]{@{}r@{}}With Disability\\ No Disability\end{tabular}  & \begin{tabular}[t]{@{}r@{}}3\\ 26\end{tabular} \\ [0.25cm] 
 \hline
 Ethnicity  & \begin{tabular}[t]{@{}r@{}}Chinese-descent\\ Indigeneous\\Other\end{tabular}  & \begin{tabular}[t]{@{}r@{}}3\\ 25\\1\end{tabular}\\[0.25cm] 
 \hline
Religion & \begin{tabular}[t]{@{}r@{}}Islam\\ Christian or Catholics\\ Hinduism or Buddhism\\ Ahmadiyya or Shia\\ Traditional Beliefs\end{tabular} & \begin{tabular}[t]{@{}r@{}}18\\ 4\\ 4\\ 2\\ 1\end{tabular}\\
\hline
Gender  & \begin{tabular}[t]{@{}r@{}}Male\\ Female\end{tabular}                                       & \begin{tabular}[t]{@{}r@{}}13\\ 16\end{tabular}\\ 
 \hline
Age  & \begin{tabular}[t]{@{}r@{}}18 - 24\\ 25 - 34\\ 35 - 44\\ 45 - 54\\55+\end{tabular}   & \begin{tabular}[t]{@{}r@{}}9\\ 8\\ 9\\ 2\\1\end{tabular}\\ 
\hline
Education & \begin{tabular}[t]{@{}r@{}}PhD Degree \\Master's Degree\\ Bachelor's Degree\\ Associate's Degree\\ High School Degree\end{tabular}                 & \begin{tabular}[t]{@{}r@{}}1\\6\\ 12\\ 2\\ 8\end{tabular}\\
\hline
Job Status   & \begin{tabular}[t]{@{}r@{}}Employed\\ College Student\\Unemployed\end{tabular}                                           & \begin{tabular}[t]{@{}r@{}}18\\ 8\\ 3\end{tabular} \\ 
\hline
Domicile   & \begin{tabular}[t]{@{}r@{}}Greater Jakarta\\ Sumatera\\Bandung Area \\ Javanese-Region\\ Other\end{tabular}                                           & \begin{tabular}[t]{@{}r@{}}10\\ 7\\4\\2\\6\end{tabular} \\ 
\hline
Presidential Vote &\begin{tabular}[t]{@{}r@{}}Candidate no. 1\\ Candidate no. 2\\ Candidate no. 3\\ Unknown or Abstain\end{tabular}                                             & \begin{tabular}[t]{@{}r@{}}9\\ 9\\ 8\\3\end{tabular} \\
\bottomrule
\end{tabular}
}
\caption{The demographic background of the 29 annotators in coarser granularity. The ethnicity demographic information that we have are more fine-grained where \textit{Indigenous} group here refers to several ethnic Indonesian groups: Java, Minang, Sunda, Bali, Dayak, Bugis, etc. with 1-2 annotators per ethnicity.}
\label{tab:annotdemo}
\end{table}

\subsection{Annotation Instrument}
To help annotators identify texts containing toxicity and/or polarization, whether explicit (e.g., direct insults) or implicit (e.g., sarcasm) \cite{krippendorff2018content}, we developed an annotation instrument. Based on  literature review and consultations with representatives from vulnerable communities, we designed a comprehensive codebook (see Appendix \ref{sec:annotguide}) that explains definitions and guide for detecting both toxic \cite[p.25--30]{Sellars_2016} and polarizing content \cite{Donohue2022,weber2021-political-polarization}. This instrument addresses the nuanced, context-dependent expressions of toxicity, an aspect that remains underexplored in prior NLP research \cite{elsherief-etal-2021-latent}.

\subsection{Data Collection and Preprocessing}
We compile our dataset by gathering Indonesian texts from multiple social media platforms. Texts from X (formerly Twitter) were collected using Brandwatch \cite{brandwatch}, while Facebook and Instagram were scraped via CrowdTangle \cite{crowdtangle}. In addition, we retrieved online news articles from CekFakta,\footnote{\url{https://cekfakta.com}} a collaborative fact-checking initiative in Indonesia. The data, spanning from September 2023 to January 2024, were scraped using a curated list of keywords indicative of hate speech targeting vulnerable groups. These keywords were derived from literature reviews, expert consultations, and focus group discussions with community representatives (see Appendix \ref{sec:scrape-keywords}). Preprocessing involved quality filtering (removing duplicates, spam, and advertisements using keyword and regex-based filters as detailed in Appendix \ref{sec:spam-keywords}) and excluding texts with fewer than four words. This processing resulted in an initial corpus of 42,846 texts, consisting of 36,550 tweets, 1,548 Facebook posts, 3,881 Instagram posts, and 867 news articles.

\subsection{Recruitment and Validation Metrics}
To ensure diverse perspectives, we recruited 28 annotators from varied demographic backgrounds, and one from our research team member (totaling 29; see Table \ref{tab:annotdemo}). Annotators were compensated at a rate of 1.14 million IDR per 1,000 texts. As a comparison, average monthly wage in Indonesia is approximately 3.5 million IDR \cite{bpsAverageWageSalary}. For quality control, we employed inter-coder reliability (ICR) metrics. Although Cohen’s Kappa is frequently used for toxicity annotations \cite{Aldreabi2023, ayele-etal-2023-exploring, vo2024exploiting}, we opted for Gwet’s AC1 due to its robustness in the presence of class imbalance \cite{Ohyama2021, Wongpakaran2013}, which suitable for our tasks.

\subsection{Annotation Process}
The annotation proceeded in two phases. During the \textbf{Training Phase}, annotators attended a comprehensive workshop on the codebook and annotated a pilot set of texts to identify toxicity (and its subtypes, such as insults, threats, profanity, identity attacks, and sexually explicit content) as well as polarized texts. Following three training sessions, annotators achieved a satisfactory Gwet’s AC1 score of 0.61 for toxicity (based on 250 sample texts), which is comparable to prior studies \cite{waseem-hovy-2016-hateful, Davidson_Warmsley_Macy_Weber_2017}, see Appendix \ref{app:preliminaries-agreement-study} for further elaboration. The inter-coder reliability for polarization was 0.37. In the \textbf{Main Annotation Phase}, annotators were assigned texts using stratified random sampling with respect to social media platform, resulting in a final annotated set of 28,477 unique texts. On average, each annotator contributed approximately 1,850 labels, with the note that some annotators completed only portions of their assignments due to the inherent mental burden of the task.

\subsection{Dataset Properties}
From 28,477 unique texts, 55.4\% were annotated by a single coder, while 44.6\% contains multiple annotations (see Appendix \ref{app:annotation-statistics} for more fine-grained statistics). As for our multi-label annotation results, Table~\ref{tab:aggregate-labelcount} summarizes the distribution of toxicity and polarization labels aggregated via majority vote, where texts with perfect disagreement were excluded. To view the label distribution of "Related to Election" and toxic types, see Appendix \ref{app:label-statistics}. 

\begin{table}[h]
\small
\begin{tabular}{c|ll|l}
\hline
\textbf{\# Coder(s)} & \multicolumn{2}{c|}{\textbf{Label}} & \textbf{\# Texts} \\ \hline
\multirow{4}{*}{$1$} & \multirow{2}{*}{Toxicity} & Toxic & 689 \\
 &  & Not Toxic & 15,059 \\ \cline{2-4} 
 & \multirow{2}{*}{Polarization} & Polarized & 2,679 \\
 &  & Not Polarized & 13,069 \\ \hline
\multirow{4}{*}{2+} & \multirow{2}{*}{Toxicity} & Toxic & 1,467 \\
 &  & Not Toxic & 9,394 \\ \cline{2-4} 
 & \multirow{2}{*}{Polarization} & Polarized & 1,132 \\
 &  & Not Polarized & 8,837 \\ \hline
\end{tabular}
\caption{Distribution of toxicity and polarization labels aggregated via majority vote.}
\label{tab:aggregate-labelcount}
\end{table}

\section{Experiment Setup and Results}
\begin{table}[h]
    \centering
    \small
    \begin{tabular}{lccc}
        \hline
        \textbf{Stats} & \textbf{Full Data} & \textbf{Toxic Exp} & \textbf{Polar Exp} \\
        \hline
        Kendall-Tau & 0.28 & 0.30 & 0.40 \\
        Cond. Prob. & 0.25 / 0.48 & 0.57 / 0.48 & 0.25 / 0.64 \\
        AUC-ROC & 0.68 / 0.60 & 0.69 / 0.71 & 0.71 / 0.59 \\
        \hline
    \end{tabular}
    \caption{Comparison of different metrics across differing split, structured as \textbf{targeting toxicity/targeting polarity} (e.g. $P(t=1|p=1) / P(p=1|t=1)$).}
    \label{tab:exp-data-stats}
\end{table}

Our dataset exhibits a strong imbalance toward non-toxic and non-polarized texts. To mitigate this, we balance each classification task separately by maintaining a 1:3 ratio between positive and negative instances. Specifically, for toxicity detection, we sample\footnote{Utilized pandas.sample (\url{https://pandas.pydata.org/docs/reference/api/pandas.DataFrame.sample.html}) with a seed of 42.} three non-toxic texts for every toxic text, resulting in 2,156 toxic texts after balancing (\textbf{Toxic Exp}). We sample our polarization detection data the same way, yielding 3,811 polarized texts in the \textbf{Polar Exp} dataset.

For annotation consistency, we employ a majority voting strategy \textbf{(AGG)}: a text is labeled as toxic or polarized if more than half of the annotators agree on the label. In most cases, this rule is strictly followed, but exceptions exist, which are discussed in relevant sections. To reduce ambiguity, we exclude texts where annotators exhibit perfect disagreement (i.e., cases where exactly half of the annotators assigned one label while the other half assigned the opposite label). Table \ref{tab:exp-data-stats} shows statistical information of the original \textbf{Full Data} and the sampled data.

\subsection{Baseline}
\begin{table*}[ht]
\centering
\setlength{\tabcolsep}{4pt} 
\renewcommand{\arraystretch}{1.2} 
\resizebox{\linewidth}{!}{
\begin{tabular}{l | c c c c c c c c}
\hline
\textbf{Metric} & \textbf{IndoBERTweet} & \textbf{NusaBERT} & \textbf{Multi-e5} & \textbf{Llama3.1-8B} & \textbf{Aya23-8B} & \textbf{SeaLLMs-7B} & \textbf{GPT-4o} & \textbf{GPT-4o-mini} \\
\hline
\multicolumn{9}{c}{\textbf{Toxicity Detection}} \\
\hline
Accuracy    & $\mathbf{.844 \,\pm\, .008}$ & $.841 \,\pm\, .005$ & $.834 \,\pm\, .007$ & $.646$ & $.750$ & $.512$ & $.829$ & $.819$ \\
Macro F1    & $\mathbf{.791 \,\pm\, .011}$ & $.779 \,\pm\, .006$ & $.776 \,\pm\, .011$ & $.631$ & $.429$ & $.505$ & $.776$ & $.775$ \\
Precision@1 & $.692 \,\pm\, .022$ & $\mathbf{.704 \,\pm\, .018}$ & $.675 \,\pm\, .015$ & $.405$ & $.000$ & $.311$ & $.649$ & $.613$ \\
Recall@1    & $.681 \,\pm\, .037$ & $.627 \,\pm\, .013$ & $.650 \,\pm\, .028$ & $\textbf{.892}$ & $.000$ & $.781$ & $.688$ & $.750$ \\
ROC AUC     & $\mathbf{.790 \,\pm\, .015}$ & $.769 \,\pm\, .006$ & $.773 \,\pm\, .013$ & --      & --      & --      & --      & --      \\
\hline
\multicolumn{9}{c}{\textbf{Polarization Detection}} \\
\hline
Accuracy    & $.801 \,\pm\, .009$ & $\mathbf{.804 \,\pm\, .010}$ & $.800 \,\pm\, .009$ & $.440$ & $.750$ & $.750$ & $.555$ & $.542$ \\
Macro F1    & $.731 \,\pm\, .013$ & $.732 \,\pm\, .016$ & $\mathbf{.735 \,\pm\, .011}$ & $.440$ & $.429$ & $.411$ & $.553$ & $.540$ \\
Precision@1 & $.608 \,\pm\, .019$ & $\mathbf{.615 \,\pm\, .019}$ & $.597 \,\pm\, .018$ & $.302$ & $.000$ & $.268$ & $.356$ & $.347$ \\
Recall@1    & $.579 \,\pm\, .027$ & $.574 \,\pm\, .038$ & $\mathbf{.612 \,\pm\, .025}$ & $.942$ & $.000$ & $.781$ & $.968$ & $.946$ \\
ROC AUC     & $.727 \,\pm\, .014$ & $.727 \,\pm\, .018$ & $\mathbf{.737 \,\pm\, .012}$ & --      & --      & --      & --      & --      \\
\hline
\end{tabular}}
\caption{Baseline model performance on toxicity and polarization detection across various models. \textbf{ROC AUC} scores are not available for LLMs.}
\label{tab:baseline-combined}
\end{table*}

We compare transformer BERT-based models \citep{koto-etal-2021-indobertweet,wang2024multilingual,wongso-etal-2025-nusabert} and Large Language Models (LLMs) \cite{openai2024gpt4ocard,aryabumi2024aya,grattafiori2024llama3herdmodels,nguyen-etal-2024-seallms}, both opaque and open-sourced, for toxicity and polarization detection. BERT-based models were evaluated using stratified 5-fold cross-validation\footnote{Utilizing scikit-learn's package (\url{https://scikit-learn.org/stable/modules/generated/sklearn.model_selection.StratifiedKFold.html}), with set seed of 42.} where we report the averaged results, whereas LLMs were evaluated in a zero-shot setup (see Appendix \ref{app:llm-2shot} for two-shot results) without any fine-tuning. All prompts are provided in Appendix \ref{app:input-prompts}.

For open-sourced models (non-GPT-4o family), we follow their respective open source licenses as available from their respective hugging-face webpage. GPT-4o usage is subject to OpenAI's API terms. Table \ref{tab:baseline-combined} shows that BERT-based models consistently outperform LLMs. IndoBERTweet \citep{koto-etal-2021-indobertweet} attains the highest average performance across both tasks, although Multi-e5 \citep{wang2024multilingual} slightly outperforms it in polarization detection.

For toxicity detection, GPT-4o and GPT-4o-mini \cite{openai2024gpt4ocard} perform comparably to neural models and to each other. However, their performance drops significantly in polarization detection, indicating polarization detection is a harder task compared to toxicity detection. Notably, Aya23-8B \cite{aryabumi2024aya} classifies all texts as non-toxic and non-polarized.

This discrepancy suggests that polarization detection is more challenging than toxicity detection. A possible explanation is that many models are explicitly trained to avoid generating toxic outputs, passively learning about toxicity detection, while polarization detection is largely neglected during training. Furthermore, toxicity detection benefits from extensive research and datasets, unlike polarization detection, leading to models struggling with the nuances of polarizing linguistic features.

Based on model performance, we conducted subsequent experiments only with IndoBERTweet and GPT-4o-mini. IndoBERTweet was selected for its strong reputation and the comparable performance of BERT-based models. GPT-4o-mini was preferred over GPT-4o due to negligible performance differences and significantly lower cost.

\subsection{Wisdom of the Crowd}
\begin{table*}[ht]
\centering
\tiny
\setlength{\tabcolsep}{4pt} 
\renewcommand{\arraystretch}{1.2} 
\begin{tabular}{l c | c c | c c | c c }
\hline
\textbf{Metric} & \textbf{Baseline} & \textbf{Single Coders} & \textbf{+Norm} & \textbf{Multiple Coders} & \textbf{+Norm} & \textbf{Multiple Coders (ANY)} & \textbf{+Norm} \\
\hline
\multicolumn{8}{c}{\textbf{Toxicity Detection}} \\
\hline
Accuracy & $\mathbf{.844 \,\pm\, .008} $ & $.831 \,\pm\, .006$ & $.824 \,\pm\, .008$ & $.827 \,\pm\, .014$ & $.835 \,\pm\, .006$ & $.828 \,\pm\, .010$ & $.780 \,\pm\, .014$ \\
Macro F1 & $\mathbf{.792 \,\pm\, .011}$ & $.746 \,\pm\, .016$ & $.728 \,\pm\, .017$ & $.785 \,\pm\, .014$ & $.782 \,\pm\, .009$ & $.786 \,\pm\, .009$ & $.709 \,\pm\, .013$ \\
Precision@1 & $.692 \,\pm\, .022$ & $\mathbf{.736 \,\pm\, .011}$ & $.736 \,\pm\, .022$ & $.628 \,\pm\, .033$ & $.666 \,\pm\, .016$ & $.627 \,\pm\, .024$ & $.560 \,\pm\, .029$ \\
Recall@1 & $.681 \,\pm\, .037$ & $.507 \,\pm\, .041$ & $.463 \,\pm\, .039$ & $.767 \,\pm\, .034$ & $.686 \,\pm\, .033$ & $\mathbf{.773 \,\pm\, .036}$ & $.573 \,\pm\, .021$ \\
ROC AUC & $.790 \,\pm\, .015$ & $.723 \,\pm\, .018$ & $.704 \,\pm\, .017$ & $.807 \,\pm\, .013$ & $.785 \,\pm\, .013$ & $\mathbf{.810 \,\pm\, .011}$ & $.711 \,\pm\, .010$ \\
\hline
\multicolumn{8}{c}{\textbf{Polarization Detection}} \\
\hline
Accuracy & $\mathbf{.801 \,\pm\, .009}$& $.796 \,\pm\, .006$ & $.793 \,\pm\, .003$ & $.787 \,\pm\, .005$ & $.781 \,\pm\, .005$ & $.767 \,\pm\, .004$ & $.778 \,\pm\, .009$ \\
Macro F1 & $.731 \,\pm\, .013$& $\mathbf{.736 \,\pm\, .008}$ & $.723 \,\pm\, .005$ & $.674 \,\pm\, .011$ & $.636 \,\pm\, .023$ & $.706 \,\pm\, .007$ & $.702 \,\pm\, .011$ \\
Precision@1 & $.608 \,\pm\, .019$& $.585 \,\pm\, .012$ & $.589 \,\pm\, .008$ & $.617 \,\pm\, .019$ & $\mathbf{.627 \,\pm\, .010}$ & $.528 \,\pm\, .008$ & $.559 \,\pm\, .022$ \\
Recall@1 & $.579 \,\pm\, .027$& $\mathbf{.637 \,\pm\, .019}$ & $.577 \,\pm\, .017$ & $.395 \,\pm\, .030$ & $.304 \,\pm\, .051$ & $.625 \,\pm\, .043$ & $.547 \,\pm\, .048$ \\
ROC AUC & $.727 \,\pm\, .014$& $\mathbf{.743 \,\pm\, .009}$ & $.721 \,\pm\, .006$ & $.657 \,\pm\, .012$ & $.622 \,\pm\, .020$ & $.719 \,\pm\, .014$ & $.701 \,\pm\, .015$ \\
\hline
\end{tabular}
\caption{Performance of each setup for the "Wisdom of the Crowd" experiment on Toxicity and Polarization tasks, with and without distribution normalization \textbf{+Norm} on the training data discussed in Section \ref{abl:wotc}. \textbf{Baseline} refers to IndoBERTweet's baseline performance.}
\label{tab:ablation-wotc}
\end{table*}

\label{exp:wotc}
Each entry of our dataset is annotated by a varied number of coders due to our annotation process (see Table \ref{tab:aggregate-labelcount}). This allows us to explore the impact of coder counts when it comes to dataset creation and how it affects model performance.

\label{exp:featural}
\begin{table*}[ht]
\centering
\small
\setlength{\tabcolsep}{6pt} 
\renewcommand{\arraystretch}{1.2} 
\begin{tabular}{l | c  c | c  c}
\hline
\textbf{Metric} & \textbf{IndoBERTweet} & \textbf{+ (AGG) Feature} & \textbf{GPT-4o-mini} & \textbf{+ (AGG) Feature} \\
\hline
\multicolumn{5}{c}{\textbf{Toxicity Detection (Using Polarization as Feature)}} \\
\hline
Accuracy    & $.844 \,\pm\, .008$ & $\mathbf{.872 \,\pm\, .008}$ & $.819$ & $.821$ \\
Macro F1    & $.791 \,\pm\, .011$ & $\mathbf{.828 \,\pm\, .011}$ & $.775$ & $.777$ \\
Precision@1 & $.692 \,\pm\, .022$ & $\mathbf{.749 \,\pm\, .019}$ & $.613$ & $.616$ \\
Recall@1    & $.681 \,\pm\, .037$ & $\mathbf{.735 \,\pm\, .033}$ & $.750$ & $.752$ \\
ROC AUC     & $.790 \,\pm\, .015$ & $\mathbf{.826 \,\pm\, .015}$ & --      & --      \\
\hline
\multicolumn{5}{c}{\textbf{Polarization Detection (Using Toxicity as Feature)}} \\
\hline
Accuracy    & $.801 \,\pm\, .009$ & $\mathbf{.820 \,\pm\, .009}$ & $.542$ & $.541$ \\
Macro F1    & $.731 \,\pm\, .013$ & $\mathbf{.757 \,\pm\, .014}$ & $.540$ & $.539$ \\
Precision@1 & $.608 \,\pm\, .019$ & $\mathbf{.645 \,\pm\, .020}$ & $.347$ & $.347$ \\
Recall@1    & $.579 \,\pm\, .027$ & $\mathbf{.622 \,\pm\, .032}$ & $.946$ & $.946$ \\
ROC AUC     & $.727 \,\pm\, .014$ & $.\mathbf{754 \,\pm\, .016}$ & --      & --      \\
\hline
\end{tabular}
\caption{Performance of IndoBERTweet and GPT-4o-mini when using cross-task features. For Toxicity Detection, polarization is used as an additional feature; for Polarization Detection, toxicity is used.}
\label{tab:featural-combined}
\end{table*}
\paragraph{Multiple-Coder Data Enhances Recall in Toxicity Detection} 
For toxicity detection, training exclusively on single-coder data yields a conservative model characterized by high precision but low recall (see Table \ref{tab:ablation-wotc}). In contrast, models trained on data annotated by multiple coders resulted in a broad-net model, achieving higher recall albeit with a reduction in precision. Notably, even though the multiple-coder subset comprises less than half of the original training data, its performance is comparable to the baseline, achieving significantly higher recall despite lower precision.

\paragraph{Maintaining Performance with Only Single-Coder Data in Polarization Detection}
For polarization detection, the effects are reversed. Training on single-coder data results in a broad-net model and a marginally higher macro F1 score relative to the baseline. Conversely, training solely on multiple-coder data produces a model with substantially lower recall and diminished performance overall. Interestingly, when we modify the labeling rule from a majority vote \textbf{(AGG)} to an \textbf{(ANY)} criterion, where an entry is labeled as polarizing if at least one annotator flags it, we obtain a model that performs only slightly below the baseline, even though it only utilizes roughly one-third of the original training data.

Although toxicity detection is inherently subjective, our findings suggest that polarization detection is even more so. In a large enough annotator pool, it is likely that at least one person will perceive a text as polarizing. This observation aligns with our dataset creation: despite efforts to standardize coder interpretations of toxicity and polarization, inter-annotator agreement for polarization is significantly lower. Consequently, models trained on polarization data with multiple annotations may struggle to generalize, as the increased annotation variability introduces more noise than informative patterns.

\subsection{Toxicity and Polarization as a Feature}

Our dataset, regardless of its designed task, contains coder annotations for both toxicity and polarization (see Table \ref{tab:exp-data-stats}). This allows us to examine the relationship between the two by using one as a feature when predicting the other. \textbf{(AGG)} features the independent variable as the average of the binary annotations, following the equation $\frac{\sum_{i=1}^{n} A_i}{n}$, where for an entry with $n$ coders, we convert the $i$\textsuperscript{th} coder's annotation $A_i$ to a binary value where "1" represents the toxic/polar text.

To integrate these values into GPT-4o-mini, we modify the input by appending: "Average [toxicity/polarization] value (ranged 0 to 1): [value]". For IndoBERTweet, we use the Indonesian translation instead. Results in Table~\ref{tab:featural-combined} show that IndoBERTweet benefits significantly from this additional information, with notable improvements in accuracy and macro F1. In contrast, GPT-4o-mini's performance remains nearly unchanged, suggesting that it does not effectively leverage the provided values. 

These findings highlight a deeper correlation between toxicity and polarization, potentially driven by the rise of toxic and polarizing texts in online discussions. The strong performance boost in IndoBERTweet suggests that jointly modeling these phenomena could be a promising direction for future research.

\subsection{Incorporating Demographic Information}
\label{sec:demo}
\begin{table*}[ht]
\centering
\small
\setlength{\tabcolsep}{6pt} 
\renewcommand{\arraystretch}{1.2} 
\resizebox{0.75\linewidth}{!}{
\begin{tabular}{l | c c c | c c c }
\hline
\multirow{2}{*}{\textbf{Metric}} & \multicolumn{3}{c|}{\textbf{IndoBERTweet}} & \multicolumn{3}{c}{\textbf{GPT-4o-mini}} \\
& \textbf{Baseline\textsuperscript{*}} & \textbf{Best} & \textbf{Worst} & \textbf{Baseline\textsuperscript{*}} & \textbf{Best} & \textbf{Worst} \\
\hline
\multicolumn{7}{c}{\textbf{Toxicity Detection}} \\
\hline
Accuracy    & $.680 \,\pm\, .007$ & $\mathbf{.832 \,\pm\, .006}$ & $.788 \,\pm\, .011$ & $.805$ & $.806$ & $.803$ \\
Macro F1    & $.405 \,\pm\, .002$ & $\mathbf{.806 \,\pm\, .004}$ & $.757 \,\pm\, .008$ & $.789$ & $.797$ & $.788$ \\
Precision@1 & $.000 \,\pm\, .000$ & $\mathbf{.744 \,\pm\, .023}$ & $.671 \,\pm\, .025$ & $.712$ & $.686$ & $.710$ \\
Recall@1    & $.000 \,\pm\, .000$ & $.728 \,\pm\, .022$ & $.671 \,\pm\, .027$ & $.753$ & $\textbf{.833}$ & $.751$ \\
ROC AUC     & $.500 \,\pm\, .000$ & $\mathbf{.805 \,\pm\, .003}$ & $.757 \,\pm\, .008$ & -- & -- & -- \\
\hline
\multicolumn{7}{c}{\textbf{Polarization Detection}} \\
\hline
Accuracy    & $.820 \,\pm\, .010$ & $\mathbf{.864 \,\pm\, .004}$ & $.836 \,\pm\, .005$ & $.530$ & $.542$ & $.527$ \\
Macro F1    & $.450 \,\pm\, .003$ & $\mathbf{.750 \,\pm\, .008}$ & $.687 \,\pm\, .009$ & $.529$ & $.540$ & $.526$ \\
Precision@1 & $.000 \,\pm\, .000$ & $\mathbf{.655 \,\pm\, .040}$ & $.562 \,\pm\, .027$ & $.349$ & $.352$ & $.345$ \\
Recall@1    & $.000 \,\pm\, .000$ & $.525 \,\pm\, .019$ & $.407 \,\pm\, .022$ & $\mathbf{.967}$ & $.962$ & $.966$ \\
ROC AUC     & $.500 \,\pm\, .000$ & $\mathbf{.732 \,\pm\, .007}$ & $.669 \,\pm\, .009$ & -- & -- & -- \\
\hline
\end{tabular}}
\caption{Performance of IndoBERTweet and GPT-4o-mini with different demographic setups. \textbf{Baseline\textsuperscript{*}} uses an exploded dataset with no demographic information. \textbf{Best} includes the coder's ethnicity, domicile, and religion. \textbf{Worst} (IndoBERTweet) includes whether the coder is disabled, while \textbf{Worst} (GPT-4o-mini) includes only the coder's age group.}
\label{tab:demographical-comparison}
\end{table*}

To incorporate demographic information into our models, we first \textbf{explode} the dataset by splitting each text annotated by $n$ coders into $n$ separate entries, each linked to a single annotator's demographic profile. Although this creates duplicate texts, each instance is uniquely associated with its coder's attributes. See Appendix \ref{app:input-prompts} for information on how we integrate demographic data into IndoBERTweet and GPT-4o-mini.

\textbf{IndoBERTweet shows a strong reliance on demographic information.} Shown in Table \ref{tab:demographical-comparison}, when trained on the exploded dataset \textit{without} demographic inputs (baseline), the model fails to distinguish between toxic or polarizing content. However, when demographic details are provided, performance improves significantly. 

The best-performing setup includes \textit{ethnicity, domicile, and religion}, achieving the highest scores across evaluation metrics. In contrast, the worst-performing setup, where the model only receives information about whether the coder is disabled, leads to the weakest results. For polarization detection, the best-performing setup also outperforms IndoBERTweet trained on the \textit{non-exploded} dataset, suggesting that demographic information contributes meaningfully to polarization detection.

\textbf{For GPT-4o-mini, however, incorporating demographic information does not significantly impact performance}. We attribute this to the rarity of these information in its training data. Though GPT-4o has been used to simulate human users, its performance has been left wanting \citep{salewski2023-incontextimpersonationrevealslarge, choi2024picle-elicitingdiversebehaviors, jiang2023evaluatinginducingpersonalitypretrained}. Compounded with the fact that this data is in Indonesian, it potentially ignores the provided demographic information. The only notable exception occurs in toxicity detection under the best setup, where recall improves substantially at the cost of lower precision, even though each of these information alone does not contribute any significant changes (see Appendix \ref{app:full-demo}). However, this does not explain why GPT-4o-mini's performance remains unchanged when provided with polarization annotations for toxicity classification and vice versa. This suggests that the model may selectively prioritize certain features over others, a behavior that warrants further investigation. Additional information on GPT-4o-mini's "persona" with respect to Indonesian identities can be found in Appendix \ref{app:gpt-persona}.

\subsection{Combining Featural and Demographic Information}

Both featural information (e.g., polarization value for toxicity classification and vice versa) and demographic information improve model performance compared to the baseline. Given this, we investigate whether combining both types of information leads to further improvements (see Appendix \ref{app:full-combined} Table \ref{tab:full-combined} for full results). Due to GPT-4o-mini's consistently unchanging performance across different demographic setups, we exclude it from this experiment, as prior results suggest that the model tends to ignores added information.

For toxicity classification, combining featural and demographic information yields the best results, achieving an F1@1 score of 0.765, significantly higher than using only featural (0.741) or demographic (0.735) information alone. Similarly, polarization classification benefits from this combination significantly, with macro F1 increasing to 0.830, compared to 0.757 (featural) and 0.750 (demographic). Notably, IndoBERTweet’s performance on polarization classification is nearly on par with toxicity classification when both information types are provided, suggesting that the model learns a shared representation for both tasks.

Overall, these results indicate that featural and demographic information complement each other, enhancing the model’s ability to detect toxic and polarizing texts more effectively than when using either information type alone.


\section{Ablation and Discussion}
\begin{table*}[h!]
\centering
\small
\setlength{\tabcolsep}{6pt} 
\renewcommand{\arraystretch}{1.2} 
\begin{tabular}{l | c | c c | c c }
\hline
\textbf{Metric} & \textbf{Baseline} & \textbf{(AGG)} & \textbf{+Pred} & \textbf{(ANY)} & \textbf{+Pred} \\
\hline
\multicolumn{6}{c}{\textbf{Toxicity}} \\
\hline
Accuracy & $.844 \,\pm\, .008$ & $\mathbf{.872 \,\pm\, .008}$ & $.869 \,\pm\, .007$ & $.867 \,\pm\, .009$ & $.834 \,\pm\, .016$ \\
Macro F1 & $.791 \,\pm\, .011$ & $\mathbf{.828 \,\pm\, .011}$ & $.824 \,\pm\, .009$ & $.823 \,\pm\, .012$ & $.722 \,\pm\, .045$ \\
Precision@1 & $.692 \,\pm\, .022$ & $.749 \,\pm\, .019$ & $.743 \,\pm\, .023$ & $.734 \,\pm\, .024$ & $\mathbf{.856 \,\pm\, .020}$ \\
Recall@1 & $.681 \,\pm\, .037$ & $.735 \,\pm\, .033$ & $.727 \,\pm\, .034$ & $\mathbf{.735 \,\pm\, .029}$ & $.406 \,\pm\, .090$ \\
ROC AUC & $.790 \,\pm\, .015$ & $\mathbf{.826 \,\pm\, .015}$ & $.821 \,\pm\, .013$ & $.823 \,\pm\, .014$ & $.691 \,\pm\, .041$ \\
\hline
\multicolumn{6}{c}{\textbf{Polarization}} \\
\hline
Accuracy & $.801 \,\pm\, .009$ & $\mathbf{.820 \,\pm\, .009}$ & $.811 \,\pm\, .005$ & $.808 \,\pm\, .009$ & $.808 \,\pm\, .005$ \\
Macro F1 & $.731 \,\pm\, .013$ & $\mathbf{.757 \,\pm\, .014}$ & $.716 \,\pm\, .018$ & $.742 \,\pm\, .014$ & $.713 \,\pm\, .020$ \\
Precision@1 & $.608 \,\pm\, .019$ & $.645 \,\pm\, .020$ & $\mathbf{.679 \,\pm\, .017}$ & $.620 \,\pm\, .019$ & $.666 \,\pm\, .014$ \\
Recall@1 & $.579 \,\pm\, .027$ & $\mathbf{.622 \,\pm\, .032}$ & $.468 \,\pm\, .052$ & $.602 \,\pm\, .031$ & $.470 \,\pm\, .064$ \\
ROC AUC & $.727 \,\pm\, .014$ & $\mathbf{.754 \,\pm\, .016}$ & $.697 \,\pm\, .020$ & $.739 \,\pm\, .015$ & $.695 \,\pm\, .024$ \\
\hline
\end{tabular}
\caption{Ablation study of Featural models on Toxicity and Polarization tasks. Performance of Predictor models are available in Appendix \ref{app:predictor-perf}. \textbf{Baseline} refers to IndoBERTweet baseline performance.}
\label{tab:ablation-featural}
\end{table*}
\subsection{How Related Are Polarization and Toxicity}
The strongest theoretical link between toxicity and polarization manifests as toxic polarization \citep{milavcic2021negative, powell2022toxic}. \citet{kolod2024polarization} define toxic polarization as "a state of intense, chronic polarization marked by high levels of loyalty to a person's ingroup and contempt or even hate for outgroups." This state deepens societal divisions, making it evident that some polarizing texts in our dataset are also toxic.

From this work, Table \ref{tab:exp-data-stats} and Experiment \ref{exp:featural} also demonstrate that toxicity can aid in predicting polarization and vice versa, thereby confirming the existence of a relationship. Table \ref{tab:exp-data-stats} further shows that using logistic regression to predict toxicity solely from the polarization label yields an AUC-ROC score exceeding 0.68 in all splits, although the results for polarization are more variable. This finding indicates that incorporating polarization as a feature for toxicity detection is more advantageous than the converse.

Notably, approximately 48\% of toxic texts during Indonesia's 2024 Presidential Election were used for polarizing purposes. Given that only 25\% of polarizing texts are toxic, our dataset suggests that Indonesia is becoming polarized at a faster rate than it is becoming toxic. This trend is particularly alarming, as Indonesia, the world's third-largest democracy, has not only seen a tenfold increase in toxicity since 2019, but also a significant portion of this toxicity may be linked to toxic polarization

\subsection{Wisdom of the Crowd on Normalized Distribution}
\label{abl:wotc}

We confirmed that the pattern observed in Result \ref{exp:wotc} is not due to distribution shifts between entries annotated by one coder and those annotated by multiple coders. This was verified by normalizing the distribution—via up-sampling or down-sampling as appropriate—to maintain a consistent class ratio of one “toxic/polarizing” entry to three “not toxic/not polarizing” entries.

Table \ref{tab:ablation-wotc} shows that, despite normalization, the original pattern persists in many cases. However, new patterns emerged in both toxicity and polarization tasks. Following normalization, both toxicity’s “Multiple Coders” condition and polarization’s “Multiple Coders (ANY)” condition achieved balanced precision@1 and recall@1, albeit with a lower macro F1 in each instance.

This validates the results in Table \ref{tab:ablation-wotc}, indicating that polarization detection may be inherently more subjective than toxicity detection. Moreover, further analysis on whether polarization detection should adhere to the same strict dataset creation protocols as toxicity detection should be done, especially given our finding that majority-based label aggregation may be counterproductive for polarization.



\subsection{Indonesian's Polarizing Identities}
Our dataset reveals identity groups characterized by high in-group agreement and significant out-group disagreement. We define these as polarizing identities, as they contribute to pronounced social divisions, measured by the gap between in-group agreement and out-group disagreement.

Based on this definition, disability emerges as the most polarizing identity in Indonesia, with a Gwet's AC1 agreement gap of 0.37 for toxicity and 0.46 for polarization. The second most polarizing identity is residence in Jakarta, as annotators from Jakarta exhibit a high Gwet's AC1 agreement gap, even compared to those from other regions within Java. The third is membership in the Gen X age group, which shows a substantial agreement gap for toxicity but a polarization agreement gap of 0 relative to other age groups. Beyond these three, most identities do not exhibit strong polarization, with education level showing the lowest agreement gap for toxicity (0.01). Full results are provided in Appendix \ref{app:aggrement-gap}.

\subsection{Non-ideal cases for Featural Experiments}
\label{abl:featural}

Experiment \ref{exp:featural} is done under an ideal situation \textbf{(AGG)}. A more realistic setup would include a simpler feature, such as utilizing a predictor or under a less-ideal format such as \textbf{(ANY)} where the independent variable is featured as a binary value following $\max(A_1, A_2, ..., A_n)$. Table \ref{tab:ablation-featural} showcases these results, showing that under \textbf{(ANY)}, the model still performs better than the baseline. However, utilizing a predictor (see Appendix \ref{app:predictor-perf}) degrades the performance massively below the baseline when it comes to both precision@1 and recall@1, with \textbf{Toxic AGG + Pred} being the only exception. 

Through ablation, we show that even under non-ideal conditions, including polarization as a feature for toxicity detection and vice versa can be helpful. Moreover, it is plausible to create a predictor for the independent variable, removing the need for human labels. However, creating a predictor through simple methods may not be adequate and is a potential area for future work.

\section{Conclusion}
We present a multi-labeled Indonesian discourse dataset of 28,477 texts annotated for toxicity, polarization, and annotator demographics—the first of its kind. Our analysis yields the following findings:

\textbf{BERT-based models outperform LLMs.} IndoBERTweet achieves the best performance across both tasks. While few-shot prompting improves LLMs in polarization detection (Appendix \ref{app:llm-2shot}), they still underperform IndoBERTweet, and few-shot setups degrade GPT-4o-mini’s toxicity detection.

\textbf{Polarization detection is more subjective than toxicity detection.} Despite extensive training, coder agreement is significantly lower for polarization, reflecting its inherent subjectivity. However, polarization labels still enhance toxicity detection, even in non-ideal conditions (Sections \ref{exp:featural}, \ref{abl:featural}).

\textbf{Demographic information aids classification but is less effective than cross-task features.} While demographics improve both toxicity and polarization detection, using polarization as a feature for toxicity (and vice versa) has a greater impact.

\textbf{Combining demographic and cross-task features further boosts performance.} This hybrid approach (Appendix \ref{app:full-combined}) improves precision@1 and recall@1, allowing polarization detection to reach performance levels comparable to toxicity detection (F1@1 = 0.748).

\textbf{GPT-4o-mini does not effectively utilize demographic information.} Likely due to its training data limitations, GPT-4o-mini ignores demographic attributes except in one setup, where recall improves at the cost of precision (Appendix \ref{app:full-demo}). Its inability to leverage polarization for toxicity detection (and vice versa) suggests selective feature prioritization, warranting further investigation (Appendix \ref{app:gpt-persona}).

\section*{Limitations} 
Our work faces several limitations, some of which reflect broader challenges in the field while others are specific to our dataset.

\paragraph{Low Inter-Coder Reliability for Polarization Detection}
Our dataset exhibits a relatively low ICR for polarization tasks; even after maintaining a 1:3 ratio of polar to non-polar texts, the ICR only increases to 0.39. Although this low score may partly be attributed to the inherent subjectivity of polarization judgments, as suggested by our "Wisdom of the Crowd" experiment, it also implies that the polarization labels may be noisy. Despite this, Table \ref{tab:exp-data-stats} showcase a moderate correlation between polarization and toxicity features exists, which proves beneficial in our cross-task experiments (Section \ref{exp:featural}).

\paragraph{Annotation Bias}
While our pool of 29 annotators is larger than that used in many non-crowdsourced toxicity datasets \citep{Davidson_Warmsley_Macy_Weber_2017, moon-etal-2020-beep, hoang-etal-2023-vihos}, Indonesia’s cultural and linguistic diversity means that this number may still be insufficient to capture all perspectives, potentially introducing bias into the annotations. Although the toxicity labels reached Gwet's AC1 scores comparable to other studies, the lower reliability for polarization suggests that additional or more diverse annotators could improve consistency.

\paragraph{Lack of Comparable Datasets}
As the first dataset to label both toxicity and polarization in this context, our work lacks a comparative baseline. This novelty makes it impossible to benchmark our models against existing resources, as they simply do not exist. The development of similar datasets in the future will be essential for contextualizing and validating our results.

\section*{Ethics Statement}
\paragraph{Balancing Risk and Benefit}
The creation of this dataset exposes annotators to potentially harmful texts. To avoid excessive mental strain, we intentionally extended the annotation duration to two and a half months. Individuals are preemptively warned and asked for consent during the initial recruitment process. Furthermore, annotators are permitted to quit the annotation process if they feel unable to proceed. We recognize the potential misuse of such datasets, which could include training models to generate more toxic and polarizing text. Yet, it’s worth noting that even without these datasets, it is alarmingly straightforward to train a model to produce toxic content, as the source of their training data, the internet, contain many of such texts. This has been demonstrated by numerous researchers who have attempted to reduce toxic output or identify vulnerabilities in large language models (refer to \citet{easy-toxic-1, easy-toxic-2}). On the other hand, the area of developing models to detect and moderate toxicity and polarizing texts, targeted at specific demographic groups is still growing, with a notable lack of available data, especially in Indonesia. Weighing these considerations, we firmly believe that the potential benefits of this type of dataset significantly outweigh the possible misuse.

\paragraph{Coder's Data Privacy}
In regards to coder's data privacy, we have ensures that all publicly available demographic information of each coder are not personally identifiable. Even with all the information combined, identifying any one of our 29 coders among the diverse 277 million  populations is improbable.

\paragraph{Responsible Use of the Dataset}
This dataset is made available solely for advancing research in detecting and moderating toxic and polarizing content, with a particular focus on Indonesian context. Users are expected to handle the data with sensitivity and ensure that any models or applications built upon it do not inadvertently promote harmful content or reinforce societal biases. The dataset should not be employed for surveillance, profiling, or any purpose that infringes on individual or community rights. Researchers and developers must implement robust privacy safeguards and conduct thorough impact assessments before deploying any systems based on this data. Any redistribution or modification of the dataset must preserve these ethical guidelines, and users are encouraged to document and share any additional measures taken to ensure its responsible use.

\section*{Acknowledgements}
Anonymized due to double-blind.

\bibliography{acl_latex}

\begin{thebibliography}{56}
\providecommand{\natexlab}[1]{#1}

\bibitem[{AJI(2024)}]{aji-hatespeech-dashboard}
AJI. 2024.
\newblock 2024 indonesian general election hate speech monitoring dashboard.
\newblock \url{https://aji.or.id/}.
\newblock Accessed June 14th, 2024.

\bibitem[{Aldreabi and Blackburn(2024)}]{Aldreabi2023}
Esraa Aldreabi and Jeremy Blackburn. 2024.
\newblock \href {https://doi.org/10.1145/3625007.3627487} {Enhancing automated hate speech detection: Addressing islamophobia and freedom of speech in online discussions}.
\newblock In \emph{Proceedings of the 2023 IEEE/ACM International Conference on Advances in Social Networks Analysis and Mining}, ASONAM '23, page 644–651, New York, NY, USA. Association for Computing Machinery.

\bibitem[{Alexandra and Satria(2023)}]{alexandra-2023}
Lina~A. Alexandra and Alif Satria. 2023.
\newblock \href {https://doi.org/10.1163/1875984x-20230005} {{Identifying Hate Speech Trends and Prevention in Indonesia: a Cross-Case Comparison}}.
\newblock \emph{Global responsibility to protect}, 15(2-3):135--176.

\bibitem[{Aryabumi et~al.(2024)Aryabumi, Dang, Talupuru, Dash, Cairuz, Lin, Venkitesh, Smith, Marchisio, Ruder, Locatelli, Kreutzer, Frosst, Blunsom, Fadaee, Üstün, and Hooker}]{aryabumi2024aya}
Viraat Aryabumi, John Dang, Dwarak Talupuru, Saurabh Dash, David Cairuz, Hangyu Lin, Bharat Venkitesh, Madeline Smith, Kelly Marchisio, Sebastian Ruder, Acyr Locatelli, Julia Kreutzer, Nick Frosst, Phil Blunsom, Marzieh Fadaee, Ahmet Üstün, and Sara Hooker. 2024.
\newblock \href {https://arxiv.org/abs/2405.15032} {Aya 23: Open weight releases to further multilingual progress}.
\newblock \emph{Preprint}, arXiv:2405.15032.

\bibitem[{Axelrod et~al.(2021)Axelrod, Daymude, and Forrest}]{compromise}
Robert Axelrod, Joshua~J. Daymude, and Stephanie Forrest. 2021.
\newblock \href {https://doi.org/10.1073/pnas.2102139118} {Preventing extreme polarization of political attitudes}.
\newblock \emph{Proceedings of the National Academy of Sciences}, 118(50):e2102139118.

\bibitem[{Ayele et~al.(2023)Ayele, Yimam, Belay, Asfaw, and Biemann}]{ayele-etal-2023-exploring}
Abinew~Ali Ayele, Seid~Muhie Yimam, Tadesse~Destaw Belay, Tesfa Asfaw, and Chris Biemann. 2023.
\newblock \href {https://aclanthology.org/2023.ranlp-1.6} {Exploring {A}mharic hate speech data collection and classification approaches}.
\newblock In \emph{Proceedings of the 14th International Conference on Recent Advances in Natural Language Processing}, pages 49--59, Varna, Bulgaria. INCOMA Ltd., Shoumen, Bulgaria.

\bibitem[{BPS-Statistics(2024)}]{bpsAverageWageSalary}
Indonesia BPS-Statistics. 2024.
\newblock \href {https://www.bps.go.id/en/statistics-table/2/MTUyMSMy/rata-rata-upah-gaji.html} {{A}verage of {N}et {W}age/{S}alary - {S}tatistical {D}ata --- bps.go.id}.

\bibitem[{Brandwatch(2021)}]{brandwatch}
Brandwatch. 2021.
\newblock Brandwatch consumer intelligence.
\newblock \url{https://www.brandwatch.com/suite/consumer-intelligence/}.

\bibitem[{Choi and Li(2024)}]{choi2024picle-elicitingdiversebehaviors}
Hyeong~Kyu Choi and Yixuan Li. 2024.
\newblock \href {https://arxiv.org/abs/2405.02501} {Picle: Eliciting diverse behaviors from large language models with persona in-context learning}.
\newblock \emph{Preprint}, arXiv:2405.02501.

\bibitem[{cjadams et~al.(2017)cjadams, Sorensen, Elliott, Dixon, McDonald, nithum, and Cukierski}]{hsclassifier-jigsaw}
cjadams, Jeffrey Sorensen, Julia Elliott, Lucas Dixon, Mark McDonald, nithum, and Will Cukierski. 2017.
\newblock \href {https://kaggle.com/competitions/jigsaw-toxic-comment-classification-challenge} {Toxic comment classification challenge}.

\bibitem[{CSIS(2022)}]{csis}
CSIS. 2022.
\newblock \href {https://hatespeech.csis.or.id/} {Hate speech dashboard}.

\bibitem[{{Data Commons}(2024)}]{datacommons-indonesia-population}
{Data Commons}. 2024.
\newblock \href {https://datacommons.org/place/country/IDN?mprop=count&popt=Person&hl=en#} {Indonesia population data}.
\newblock Accessed: 2024-12-19.

\bibitem[{Davidson et~al.(2017)Davidson, Warmsley, Macy, and Weber}]{Davidson_Warmsley_Macy_Weber_2017}
Thomas Davidson, Dana Warmsley, Michael Macy, and Ingmar Weber. 2017.
\newblock \href {https://doi.org/10.1609/icwsm.v11i1.14955} {Automated hate speech detection and the problem of offensive language}.
\newblock \emph{Proceedings of the International AAAI Conference on Web and Social Media}, 11(1):512--515.

\bibitem[{DiMaggio et~al.(1996)DiMaggio, Evans, and Bryson}]{dimaggio1996polarized}
Paul DiMaggio, John Evans, and Bethany Bryson. 1996.
\newblock Have american's social attitudes become more polarized?
\newblock \emph{American Journal of Sociology}, 102(3):690--755.

\bibitem[{Donohue and Hamilton(2022)}]{Donohue2022}
William Donohue and Mark Hamilton. 2022.
\newblock \href {https://doi.org/10.4324/9780367823658-14} {\emph{A Framework for Understanding Polarizing Language}}, 1 edition.
\newblock Routledge.

\bibitem[{ElSherief et~al.(2021)ElSherief, Ziems, Muchlinski, Anupindi, Seybolt, De~Choudhury, and Yang}]{elsherief-etal-2021-latent}
Mai ElSherief, Caleb Ziems, David Muchlinski, Vaishnavi Anupindi, Jordyn Seybolt, Munmun De~Choudhury, and Diyi Yang. 2021.
\newblock \href {https://doi.org/10.18653/v1/2021.emnlp-main.29} {Latent hatred: A benchmark for understanding implicit hate speech}.
\newblock In \emph{Proceedings of the 2021 Conference on Empirical Methods in Natural Language Processing}, pages 345--363, Online and Punta Cana, Dominican Republic. Association for Computational Linguistics.

\bibitem[{Gehman et~al.(2020)Gehman, Gururangan, Sap, Choi, and Smith}]{easy-toxic-1}
Samuel Gehman, Suchin Gururangan, Maarten Sap, Yejin Choi, and Noah~A. Smith. 2020.
\newblock \href {https://arxiv.org/abs/2009.11462} {Realtoxicityprompts: Evaluating neural toxic degeneration in language models}.
\newblock \emph{Preprint}, arXiv:2009.11462.

\bibitem[{Grattafiori et~al.(2024)Grattafiori, Dubey, Jauhri, Pandey, Kadian, Al-Dahle, Letman, Mathur, Schelten, Vaughan, Yang, Fan, Goyal, Hartshorn, Yang, Mitra, Sravankumar, Korenev, Hinsvark, Rao, Zhang, Rodriguez, Gregerson, Spataru, Roziere, Biron, Tang, Chern, Caucheteux, Nayak, Bi, Marra, McConnell, Keller, Touret, Wu, Wong, Ferrer, Nikolaidis, Allonsius, Song, Pintz, Livshits, Wyatt, Esiobu, Choudhary, Mahajan, and et~al}]{grattafiori2024llama3herdmodels}
Aaron Grattafiori, Abhimanyu Dubey, Abhinav Jauhri, Abhinav Pandey, Abhishek Kadian, Ahmad Al-Dahle, Aiesha Letman, Akhil Mathur, Alan Schelten, Alex Vaughan, Amy Yang, Angela Fan, Anirudh Goyal, Anthony Hartshorn, Aobo Yang, Archi Mitra, Archie Sravankumar, Artem Korenev, Arthur Hinsvark, Arun Rao, Aston Zhang, Aurelien Rodriguez, Austen Gregerson, Ava Spataru, Baptiste Roziere, Bethany Biron, Binh Tang, Bobbie Chern, Charlotte Caucheteux, Chaya Nayak, Chloe Bi, Chris Marra, Chris McConnell, Christian Keller, Christophe Touret, Chunyang Wu, Corinne Wong, Cristian~Canton Ferrer, Cyrus Nikolaidis, Damien Allonsius, Daniel Song, Danielle Pintz, Danny Livshits, Danny Wyatt, David Esiobu, Dhruv Choudhary, Dhruv Mahajan, and Diego Garcia-Olano et~al. 2024.
\newblock \href {https://arxiv.org/abs/2407.21783} {The llama 3 herd of models}.
\newblock \emph{Preprint}, arXiv:2407.21783.

\bibitem[{Haber et~al.(2023)Haber, Vidgen, Chapman, Agarwal, Lee, Yap, and R{\"o}ttger}]{haber-etal-2023-improving}
Janosch Haber, Bertie Vidgen, Matthew Chapman, Vibhor Agarwal, Roy Ka-Wei Lee, Yong~Keong Yap, and Paul R{\"o}ttger. 2023.
\newblock \href {https://doi.org/10.18653/v1/2023.acl-long.711} {Improving the detection of multilingual online attacks with rich social media data from {S}ingapore}.
\newblock In \emph{Proceedings of the 61st Annual Meeting of the Association for Computational Linguistics (Volume 1: Long Papers)}, pages 12705--12721, Toronto, Canada. Association for Computational Linguistics.

\bibitem[{Hoang et~al.(2023)Hoang, Luu, Tran, Nguyen, and Nguyen}]{hoang-etal-2023-vihos}
Phu~Gia Hoang, Canh~Duc Luu, Khanh~Quoc Tran, Kiet~Van Nguyen, and Ngan Luu-Thuy Nguyen. 2023.
\newblock \href {https://doi.org/10.18653/v1/2023.eacl-main.47} {{V}i{HOS}: Hate speech spans detection for {V}ietnamese}.
\newblock In \emph{Proceedings of the 17th Conference of the European Chapter of the Association for Computational Linguistics}, pages 652--669, Dubrovnik, Croatia. Association for Computational Linguistics.

\bibitem[{HOBOLT et~al.(2024)HOBOLT, LAWALL, and TILLEY}]{echo-chamber}
SARA~B. HOBOLT, KATHARINA LAWALL, and JAMES TILLEY. 2024.
\newblock \href {https://doi.org/10.1017/S0003055423001211} {The polarizing effect of partisan echo chambers}.
\newblock \emph{American Political Science Review}, 118(3):1464–1479.

\bibitem[{Jiang et~al.(2023)Jiang, Xu, Zhu, Han, Zhang, and Zhu}]{jiang2023evaluatinginducingpersonalitypretrained}
Guangyuan Jiang, Manjie Xu, Song-Chun Zhu, Wenjuan Han, Chi Zhang, and Yixin Zhu. 2023.
\newblock \href {https://arxiv.org/abs/2206.07550} {Evaluating and inducing personality in pre-trained language models}.
\newblock \emph{Preprint}, arXiv:2206.07550.

\bibitem[{john~a. powell(2022)}]{powell2022toxic}
john~a. powell. 2022.
\newblock \href {https://doi.org/10.24926/25730037.645} {Overcoming toxic polarization: Lessons in effective bridging}.
\newblock \emph{Law \& Inequality}, 40(2):247.

\bibitem[{KhudaBukhsh et~al.(2021)KhudaBukhsh, Sarkar, Kamlet, and Mitchell}]{khudabukhsh2021we}
Ashiqur~R KhudaBukhsh, Rupak Sarkar, Mark~S Kamlet, and Tom Mitchell. 2021.
\newblock We don't speak the same language: Interpreting polarization through machine translation.
\newblock In \emph{Proceedings of the AAAI Conference on Artificial Intelligence}, volume~35, pages 14893--14901.

\bibitem[{Kolod et~al.(2024)Kolod, Freeman-Carroll, Glover, Gurdal, Kwintner, Lysa, Moses, Podder, Raisi, Resnizky, Yanchyshyn, Zhilinskaya, and Zimmermann}]{kolod2024polarization}
Sue Kolod, Nancy Freeman-Carroll, William Glover, Cemile~Serin Gurdal, Michelle Kwintner, Tamara Lysa, Lizbeth Moses, Jhelum Podder, Hossein Raisi, Silvia Resnizky, Gordon Yanchyshyn, Alena Zhilinskaya, and Heloisa Zimmermann. 2024.
\newblock \href {https://www.ipa.world/IPA/IPA_DOCS/PDFDocuments/2024-08-IPA-Thinkinglabs-Kolod.pdf} {Thinking labs: Political polarization and social identity}.
\newblock Accessed: 2024-12-19.

\bibitem[{Koto et~al.(2021)Koto, Lau, and Baldwin}]{koto-etal-2021-indobertweet}
Fajri Koto, Jey~Han Lau, and Timothy Baldwin. 2021.
\newblock \href {https://doi.org/10.18653/v1/2021.emnlp-main.833} {{I}ndo{BERT}weet: A pretrained language model for {I}ndonesian {T}witter with effective domain-specific vocabulary initialization}.
\newblock In \emph{Proceedings of the 2021 Conference on Empirical Methods in Natural Language Processing}, pages 10660--10668, Online and Punta Cana, Dominican Republic. Association for Computational Linguistics.

\bibitem[{Krippendorff(2018)}]{krippendorff2018content}
Klaus Krippendorff. 2018.
\newblock \emph{Content analysis: An introduction to its methodology}.
\newblock Sage publications.

\bibitem[{Kumar et~al.(2021)Kumar, Kelley, Consolvo, Mason, Bursztein, Durumeric, Thomas, and Bailey}]{kumar-2021}
Deepak Kumar, Patrick~Gage Kelley, Sunny Consolvo, Joshua Mason, Elie Bursztein, Zakir Durumeric, Kurt Thomas, and Michael Bailey. 2021.
\newblock \href {https://www.usenix.org/conference/soups2021/presentation/kumar} {Designing toxic content classification for a diversity of perspectives}.
\newblock In \emph{Seventeenth Symposium on Usable Privacy and Security (SOUPS 2021)}, pages 299--318. USENIX Association.

\bibitem[{Lima et~al.(2024)Lima, Pagano, and da~Silva}]{lima-etal-2024-toxic}
Luiz Henrique~Quevedo Lima, Adriana~Silvina Pagano, and Ana Paula~Couto da~Silva. 2024.
\newblock \href {https://aclanthology.org/2024.propor-1.48/} {Toxic content detection in online social networks: a new dataset from {B}razilian {R}eddit communities}.
\newblock In \emph{Proceedings of the 16th International Conference on Computational Processing of Portuguese - Vol. 1}, pages 472--482, Santiago de Compostela, Galicia/Spain. Association for Computational Lingustics.

\bibitem[{Löfgren~Nilsson and Örnebring(2020)}]{toxicity-threatens-journalism}
Monica Löfgren~Nilsson and Henrik Örnebring. 2020.
\newblock \href {https://doi.org/10.4324/9780429462030-22} {Journalism under threat}.
\newblock \emph{Taylor and Francis}, pages 217--227.

\bibitem[{McCoy and Somer(2018)}]{mccoy2018pernicious}
Jennifer McCoy and Murat Somer. 2018.
\newblock Toward a theory of pernicious polarization and how it harms democracies: Comparative evidence and possible remedies.
\newblock \emph{The ANNALS of the American Academy of Political and Social Science}, 681(1):234--271.

\bibitem[{Midtbøen(2018)}]{toxicity-self-censorship}
Arnfinn~H Midtbøen. 2018.
\newblock \href {https://doi.org/10.1177/1468796816684149} {The making and unmaking of ethnic boundaries in the public sphere: The case of norway}.
\newblock \emph{Ethnicities}, 18(3):344--362.

\bibitem[{Mila{\v{c}}i{\'c}(2021)}]{milavcic2021negative}
Filip Mila{\v{c}}i{\'c}. 2021.
\newblock The negative impact of polarization on democracy.
\newblock \emph{Friedrich--Ebert-Stiftung. https://library. fes. de/pdf-files/bueros/wien/18175. pdf}.

\bibitem[{Moon et~al.(2020)Moon, Cho, and Lee}]{moon-etal-2020-beep}
Jihyung Moon, Won~Ik Cho, and Junbum Lee. 2020.
\newblock \href {https://doi.org/10.18653/v1/2020.socialnlp-1.4} {{BEEP}! {K}orean corpus of online news comments for toxic speech detection}.
\newblock In \emph{Proceedings of the Eighth International Workshop on Natural Language Processing for Social Media}, pages 25--31, Online. Association for Computational Linguistics.

\bibitem[{Nguyen et~al.(2024)Nguyen, Zhang, Li, Aljunied, Hu, Shen, Chia, Li, Wang, Tan, Cheng, Chen, Deng, Yang, Liu, Zhang, and Bing}]{nguyen-etal-2024-seallms}
Xuan-Phi Nguyen, Wenxuan Zhang, Xin Li, Mahani Aljunied, Zhiqiang Hu, Chenhui Shen, Yew~Ken Chia, Xingxuan Li, Jianyu Wang, Qingyu Tan, Liying Cheng, Guanzheng Chen, Yue Deng, Sen Yang, Chaoqun Liu, Hang Zhang, and Lidong Bing. 2024.
\newblock \href {https://doi.org/10.18653/v1/2024.acl-demos.28} {{S}ea{LLM}s - large language models for {S}outheast {A}sia}.
\newblock In \emph{Proceedings of the 62nd Annual Meeting of the Association for Computational Linguistics (Volume 3: System Demonstrations)}, pages 294--304, Bangkok, Thailand. Association for Computational Linguistics.

\bibitem[{Ohyama(2021)}]{Ohyama2021}
Tetsuji Ohyama. 2021.
\newblock \href {https://doi.org/10.1080/03610926.2019.1708397} {Statistical inference of gwet’s ac1 coefficient for multiple raters and binary outcomes}.
\newblock \emph{Communications in Statistics - Theory and Methods}, 50(15):3564--3572.

\bibitem[{OpenAI et~al.(2024)OpenAI, Hurst, Lerer, Goucher, Perelman, Ramesh, Clark, Ostrow, Welihinda, Hayes, Radford, Mądry, Baker-Whitcomb, Beutel, Borzunov, Carney, Chow, Kirillov, Nichol, Paino, Renzin, Passos, Kirillov, Christakis, Conneau, Kamali, Jabri, Moyer, Tam, Crookes, Tootoochian, and et~al}]{openai2024gpt4ocard}
OpenAI, Aaron Hurst, Adam Lerer, Adam~P. Goucher, Adam Perelman, Aditya Ramesh, Aidan Clark, AJ~Ostrow, Akila Welihinda, Alan Hayes, Alec Radford, Aleksander Mądry, Alex Baker-Whitcomb, Alex Beutel, Alex Borzunov, Alex Carney, Alex Chow, Alex Kirillov, Alex Nichol, Alex Paino, Alex Renzin, Alex~Tachard Passos, Alexander Kirillov, Alexi Christakis, Alexis Conneau, Ali Kamali, Allan Jabri, Allison Moyer, Allison Tam, Amadou Crookes, Amin Tootoochian, and Amin~Tootoonchian et~al. 2024.
\newblock \href {https://arxiv.org/abs/2410.21276} {Gpt-4o system card}.
\newblock \emph{Preprint}, arXiv:2410.21276.

\bibitem[{Putnam(2007)}]{putnam-2007}
Robert Putnam. 2007.
\newblock \href {https://doi.org/10.1111/j.1467-9477.2007.00176.x} {E pluribus unum: Diversity and community in the twenty-first century – the 2006 johan skytte prize lecture}.
\newblock \emph{Scandinavian Political Studies}, 30:137 -- 174.

\bibitem[{Romero-Rodríguez et~al.(2023)Romero-Rodríguez, Castillo-Abdul, and Cuesta-Valiño}]{PaG6663}
Luis Romero-Rodríguez, Bárbara Castillo-Abdul, and Pedro Cuesta-Valiño. 2023.
\newblock \href {https://doi.org/10.17645/pag.v11i2.6663} {The process of the transfer of hate speech to demonization and social polarization}.
\newblock \emph{Politics and Governance}, 11(2):109--113.

\bibitem[{Salewski et~al.(2023)Salewski, Alaniz, Rio-Torto, Schulz, and Akata}]{salewski2023-incontextimpersonationrevealslarge}
Leonard Salewski, Stephan Alaniz, Isabel Rio-Torto, Eric Schulz, and Zeynep Akata. 2023.
\newblock \href {https://arxiv.org/abs/2305.14930} {In-context impersonation reveals large language models' strengths and biases}.
\newblock \emph{Preprint}, arXiv:2305.14930.

\bibitem[{Schweighofer(2018)}]{schweighofer2018polarization}
Simon Schweighofer. 2018.
\newblock \emph{Affective, Cognitive and Social Identity Related Factors of Political Polarization}.
\newblock ETH Zurich, Salzburg.

\bibitem[{Sellars(2016)}]{Sellars_2016}
Andrew Sellars. 2016.
\newblock \href {https://doi.org/10.2139/ssrn.2882244} {Defining hate speech}.
\newblock \emph{Social Science Research Network}.

\bibitem[{Sinno et~al.(2022)Sinno, Oviedo, Atwell, Alikhani, and Li}]{sinno2022political}
Barea Sinno, Bernardo Oviedo, Katherine Atwell, Malihe Alikhani, and Junyi~Jessy Li. 2022.
\newblock Political ideology and polarization: A multi-dimensional approach.
\newblock In \emph{Proceedings of the 2022 Conference of the North American Chapter of the Association for Computational Linguistics: Human Language Technologies}, pages 231--243.

\bibitem[{Szwoch et~al.(2022)Szwoch, Staszkow, Rzepka, and Araki}]{szwoch-etal-2022-creation}
Joanna Szwoch, Mateusz Staszkow, Rafal Rzepka, and Kenji Araki. 2022.
\newblock \href {https://aclanthology.org/2022.politicalnlp-1.12/} {Creation of {P}olish online news corpus for political polarization studies}.
\newblock In \emph{Proceedings of the LREC 2022 workshop on Natural Language Processing for Political Sciences}, pages 86--90, Marseille, France. European Language Resources Association.

\bibitem[{Team(2024)}]{crowdtangle}
CrowdTangle Team. 2024.
\newblock Crowdtangle.
\newblock Facebook, Menlo Park, Califormnia, United States.
\newblock 1816403,1824912.

\bibitem[{Turner and Smaldino(2018)}]{extremism}
Matthew~A. Turner and Paul~E. Smaldino. 2018.
\newblock \href {https://doi.org/10.1155/2018/2740959} {Paths to polarization: How extreme views, miscommunication, and random chance drive opinion dynamics}.
\newblock \emph{Complexity}, 2018(1):2740959.

\bibitem[{Vasist et~al.(2024)Vasist, Chatterjee, and Krishnan}]{Vasist2024}
Pramukh~Nanjundaswamy Vasist, Debashis Chatterjee, and Satish Krishnan. 2024.
\newblock \href {https://doi.org/10.1007/s10796-023-10390-w} {The polarizing impact of political disinformation and hate speech: A cross-country configural narrative}.
\newblock \emph{Information Systems Frontiers}, 26(2):663--688.

\bibitem[{Vo et~al.(2024)Vo, Huynh, Luu, and Do}]{vo2024exploiting}
Cuong~Nhat Vo, Khanh~Bao Huynh, Son~T. Luu, and Trong-Hop Do. 2024.
\newblock \href {https://arxiv.org/abs/2404.19252} {Exploiting hatred by targets for hate speech detection on vietnamese social media texts}.
\newblock \emph{Preprint}, arXiv:2404.19252.

\bibitem[{Vorakitphan et~al.(2020)Vorakitphan, Guerini, Cabrio, and Villata}]{vorakitphan-etal-2020-regrexit}
Vorakit Vorakitphan, Marco Guerini, Elena Cabrio, and Serena Villata. 2020.
\newblock \href {https://doi.org/10.18653/v1/2020.coling-main.19} {Regrexit or not regrexit: Aspect-based sentiment analysis in polarized contexts}.
\newblock In \emph{Proceedings of the 28th International Conference on Computational Linguistics}, pages 219--224, Barcelona, Spain (Online). International Committee on Computational Linguistics.

\bibitem[{Wang et~al.(2024)Wang, Yang, Huang, Yang, Majumder, and Wei}]{wang2024multilingual}
Liang Wang, Nan Yang, Xiaolong Huang, Linjun Yang, Rangan Majumder, and Furu Wei. 2024.
\newblock Multilingual e5 text embeddings: A technical report.
\newblock \emph{arXiv preprint arXiv:2402.05672}.

\bibitem[{Waseem and Hovy(2016)}]{waseem-hovy-2016-hateful}
Zeerak Waseem and Dirk Hovy. 2016.
\newblock \href {https://doi.org/10.18653/v1/N16-2013} {Hateful symbols or hateful people? predictive features for hate speech detection on {T}witter}.
\newblock In \emph{Proceedings of the {NAACL} Student Research Workshop}, pages 88--93, San Diego, California. Association for Computational Linguistics.

\bibitem[{Weber et~al.(2021)Weber, Hydock, Ding, Gardner, Jacob, Mandel, Sprott, and Steenburg}]{weber2021-political-polarization}
T.J. Weber, Chris Hydock, William Ding, Meryl Gardner, Pradeep Jacob, Naomi Mandel, David~E. Sprott, and Eric~Van Steenburg. 2021.
\newblock \href {https://doi.org/10.1177/0743915621991103} {Political polarization: Challenges, opportunities, and hope for consumer welfare, marketers, and public policy}.
\newblock \emph{Journal of Public Policy \& Marketing}, 40(2):184--205.

\bibitem[{Wen et~al.(2023)Wen, Ke, Sun, Zhang, Li, Bai, and Huang}]{easy-toxic-2}
Jiaxin Wen, Pei Ke, Hao Sun, Zhexin Zhang, Chengfei Li, Jinfeng Bai, and Minlie Huang. 2023.
\newblock \href {https://doi.org/10.18653/v1/2023.emnlp-main.84} {Unveiling the implicit toxicity in large language models}.
\newblock In \emph{Proceedings of the 2023 Conference on Empirical Methods in Natural Language Processing}, pages 1322--1338, Singapore. Association for Computational Linguistics.

\bibitem[{Williams et~al.(2019)Williams, Burnap, Javed, Liu, and Ozalp}]{hatespeech-williams-2019}
Matthew~L Williams, Pete Burnap, Amir Javed, Han Liu, and Sefa Ozalp. 2019.
\newblock \href {https://doi.org/10.1093/bjc/azz049} {{Hate in the Machine: Anti-Black and Anti-Muslim Social Media Posts as Predictors of Offline Racially and Religiously Aggravated Crime}}.
\newblock \emph{The British Journal of Criminology}, 60(1):93--117.

\bibitem[{Wongpakaran et~al.(2013)Wongpakaran, Wongpakaran, Wedding, and Gwet}]{Wongpakaran2013}
Nahathai Wongpakaran, Tinakon Wongpakaran, Danny Wedding, and Kilem~L Gwet. 2013.
\newblock \href {https://doi.org/10.1186/1471-2288-13-61} {A comparison of cohen’s kappa and gwet’s ac1 when calculating inter-rater reliability coefficients: a study conducted with personality disorder samples}.
\newblock \emph{BMC Medical Research Methodology}, 13(1).

\bibitem[{Wongso et~al.(2025)Wongso, Setiawan, Limcorn, and Joyoadikusumo}]{wongso-etal-2025-nusabert}
Wilson Wongso, David~Samuel Setiawan, Steven Limcorn, and Ananto Joyoadikusumo. 2025.
\newblock \href {https://aclanthology.org/2025.sealp-1.2/} {{N}usa{BERT}: Teaching {I}ndo{BERT} to be multilingual and multicultural}.
\newblock In \emph{Proceedings of the Second Workshop in South East Asian Language Processing}, pages 10--26, Online. Association for Computational Linguistics.

\end{thebibliography}

\onecolumn
\newpage

\appendix

\section{Data Scraping and Preprocessing}
\subsection{Keywords Used for Scraping}
\label{sec:scrape-keywords}
cina, china, tionghoa, chinese, cokin, cindo, chindo, shia, syiah, syia, ahmadiyya, ahmadiyah, ahmadiya, ahmadiyyah, transgender, queer, bisexual, bisex, gay, lesbian, lesbong, gangguan jiwa, gangguan mental, lgbt, eljibiti, lgbtq+, lghdtv+, katolik, khatolik, kristen, kris10, kr1st3n, buta, tuli, bisu, budek, conge, idiot, autis, orang gila, orgil, gila, gendut, cacat, odgj, zionis, israel, jewish, jew, yahudi, joo, anti-christ, anti kristus, anti christ, netanyahu, setanyahu, bangsa pengecut, is ra hell, rohingya, pengungsi, imigran, sakit jiwa, tuna netra, tuna rungu, sinting.

\subsection{Keywords Used for Removing Spam Texts}
\label{sec:spam-keywords}
\#openBO, \#partnerpasutri, \#JudiOnline, Slot Gacor, \#pijat[a-z]+, \#gigolo[a-z]+, \#pasutri[a-z]+, pijit sensual, \#sangekberat, \#viralmesum, "privasi terjamin 100\%", privasi 100\%, ready open, ready partner, ready pijat, ready sayang, \#sangeberat, obat herbal, no minus, new produk

\section{Annotation Guidelines}
\label{sec:annotguide}
\subsection{Toxic Messages Definition}
\textbf{Toxic comments}\quad is a post, text, or comment that is harsh, impolite, or nonsensical, causing you to become silent and unresponsive, or that is filled with hatred and aggression, provoking feelings of disgust, anger, sadness, or humiliation, making you want to leave the discussion or give up sharing your opinion.

\textbf{Profanity or Obscenity}\quad The message / sentence on social media posts contains offensive, indecent, or inappropriate in a way that goes against accepted social norms. It often involves explicit or vulgar language, graphic content, or inappropriate references. Essentially, it's a message that is likely to be considered offensive or objectionable by most people.

\textbf{Threat / Incitement to Violence}\quad The message / sentence on social media posts conveys an intent to cause harm, danger, or significant distress to an individual or a group. It often includes explicit or implicit threats of violence, physical harm, intimidation, or any action that creates a sense of fear or apprehension.

\textbf{Insults}\quad The message / sentence on social media posts contains offensive, disrespectful, or scornful language with the intention of belittling, offending, or hurting the feelings.

\textbf{Identity Attack}\quad The message / sentence on social media posts deliberately targets and undermines a person's sense of self, identity, or personal characteristics. This can include derogatory comments, or harmful statements aimed at aspects such as one's race, gender, sexual orientation, religion, appearance, or other defining attributes.

\textbf{Sexually Explicit}\quad The message / sentence on social media posts contains explicit and detailed descriptions or discussions of sexual activities, body parts, or other related content.

\subsection{Polarizing Messages Definition}
\textbf{Polarizing Messages} is a post, text, or comment with purpose to promote conflict between two or more groups of people, often by presenting a highly biased or extreme perspective on a particular topic. A polarizing messages are designed to provoke strong reactions and attract individuals with similar beliefs, while potentially alienating or opposing those with differing perspectives.

\subsection{Manual Annotation}
Table \ref{tab:annotation-form} shows the list of questions that was asked to annotators for the annotation tasks.
\begin{table}[h!]
\centering
\resizebox{\columnwidth}{!}{
\begin{tabular}{|cll|}
\hline
\multicolumn{3}{|c|}{\textbf{Annotation Form}} \\ \hline
\multicolumn{1}{|c|}{Q1} & \multicolumn{1}{l|}{\textbf{Does this text appear to be random spam or lack context?}} & \begin{tabular}[c]{@{}l@{}} \textbullet\ Yes\\ \textbullet\ No\end{tabular} \\ \hline
\multicolumn{1}{|c|}{Q2} & \multicolumn{1}{l|}{\textbf{Does this text related to Indonesian 2024 General Election?}} & \begin{tabular}[c]{@{}l@{}}\textbullet\ Yes\\ \textbullet\ No\end{tabular} \\ \hline
\multicolumn{1}{|c|}{Q3} & \multicolumn{1}{l|}{\textbf{Does this text polarized?} } & \begin{tabular}[c]{@{}l@{}}\textbullet\ Yes\\ \textbullet\ No\end{tabular} \\ \hline
\multicolumn{1}{|c|}{Q4} & \multicolumn{1}{l|}{\begin{tabularx}{\linewidth}{X}
     \textbf{Does this text contain toxicity? }\\
     \textit{Note}: Irrelevant toxicity or hate speech includes hate speech that is meant as a joke among friends or is not considered hate speech by the recipient. Thus, it will be coded as "No".
\end{tabularx}} & \begin{tabular}[c]{@{}l@{}}\textbullet\ Yes\\ \textbullet\ No\end{tabular} \\ \hline
\multicolumn{1}{|c|}{Q5} & \multicolumn{1}{l|}{\begin{tabularx}{\linewidth}{X}
     \textbf{What is the type of toxicity?}\\
     \textit{Note:} Checkmark one or more types. Consider the following sentences as an example: \textit{“PDIP Provokasi Massa pendukungnya geruduk kediaman Anies”} (\textit{"Political party PDIP incites their supporters to storm Anies' residence"}). This headline should be coded as both threat and incitement to violence.\\
\end{tabularx}} &   \begin{tabular}[c]{@{}l@{}}$\square$\ Insults\\ $\square$\ Threat
\\ $\square$\ Profanity \\ $\square$\ Identity Attack \\ $\square$\ Sexually Explicit
\end{tabular} \\\hline

\end{tabular}}
\caption{List of questions given to annotators for every text.}
\label{tab:annotation-form}
\end{table}

\newpage
\section{Example of Toxic, Politically Polarizing, and Both}
\label{app:example-5sample}

\begin{figure}[h]
    \centering
    \large
    \includegraphics[width=0.95\linewidth]{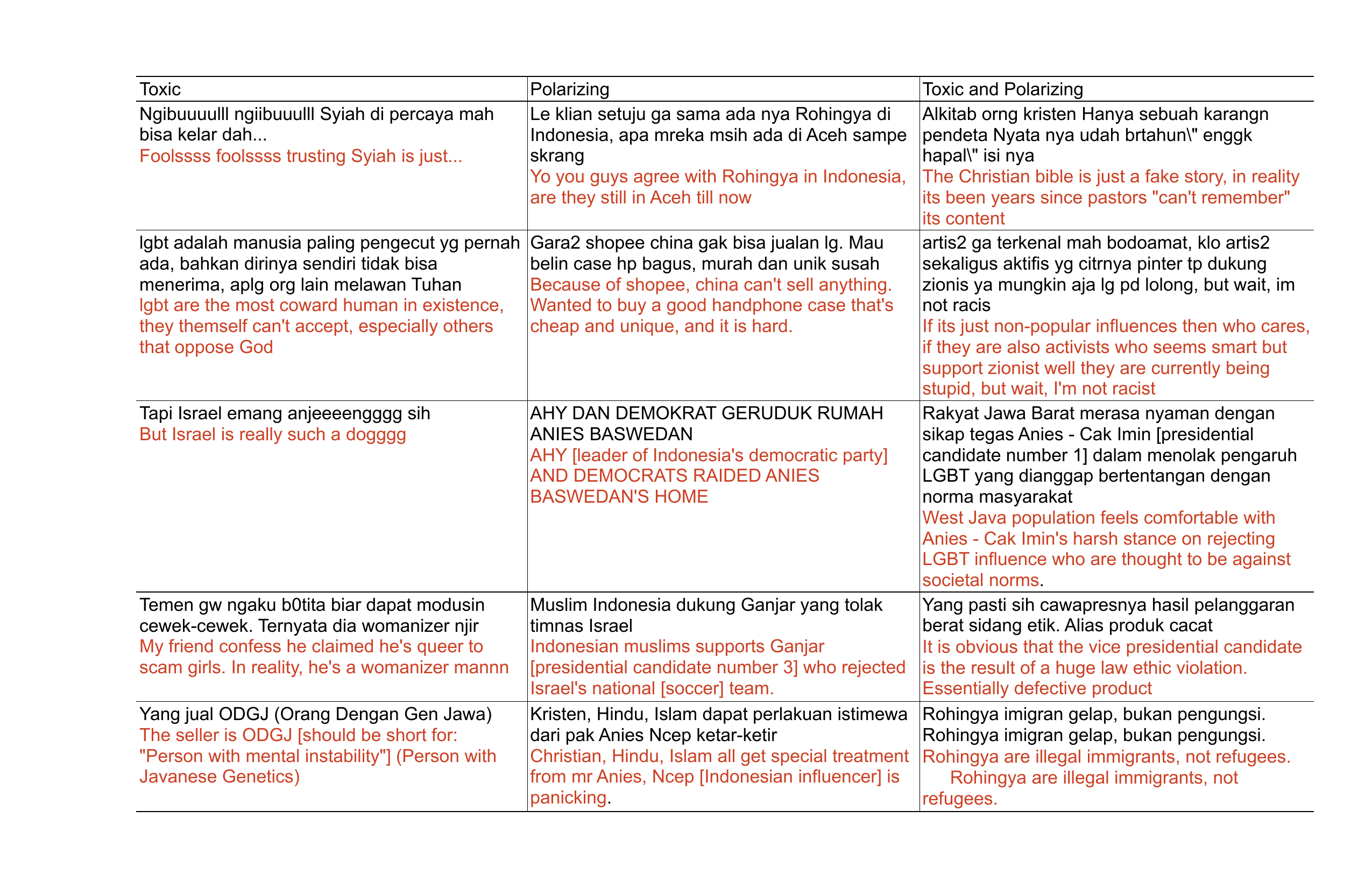}
    \caption{Samples of Toxic, Polarizing, alongside both Toxic and Polarizing texts.}
    \label{fig:text samples}
\end{figure}

\newpage
\section{Dataset Comparison}
\label{app:dataset-comparison}
\begin{table*}[h]
\centering
\begin{tabular}{lccccccc}
\toprule
\textbf{Dataset} & \textbf{Entry} & \textbf{Language} &  \textbf{Toxic} & \textbf{Polar} & \textbf{Identity} \\
\midrule
\textbf{Ours}  & \textbf{28K} & \textbf{Indonesian} & \cmark & \cmark & \cmark \\
\citet{Davidson_Warmsley_Macy_Weber_2017} & 25K & English & \cmark & \xmark & \xmark \\
\citet{moon-etal-2020-beep} & 9K & Korean & \cmark & \xmark & \cmark \\
\citet{vorakitphan-etal-2020-regrexit} & 67K\textsuperscript{a} & English & \xmark & \cmark & \xmark \\
\citet{kumar-2021} & 107K & English & \cmark & \xmark & \cmark \\
\citet{sinno2022political} & 1K\textsuperscript{p} & English & \xmark & \cmark & \xmark \\
\citet{szwoch-etal-2022-creation} & 16k\textsuperscript{a} & Polish & \xmark & \cmark & \xmark \\
\citet{hoang-etal-2023-vihos} & 11K & Vietnamese & \cmark & \xmark & \cmark \\
\citet{lima-etal-2024-toxic} & 6M* & Brazilian Portuguese & \cmark & \xmark & \xmark \\
\bottomrule
\end{tabular}
\caption{Comparison of Datasets. Unless specified, entry counts are sentence/comment level. Superscript \textsuperscript{a} and \textsuperscript{p} denotes \textbf{"Article"} and \textbf{"Paragraphs"} level data respectively. \citet{lima-etal-2024-toxic} utilizes Perspective API \citep{hsclassifier-jigsaw} for automatic labeling.}
\label{tab:dataset_comparison}
\end{table*}

\section{Notes on Agreement Score}
\label{app:preliminaries-agreement-study}
To establish a clearer understanding of what considered as a \textit{good ICR score}, we conducted literature review on several sources. However, due to variations in measurement methods and to ensure a more robust comparison, we recalculated the ICR metric internally. However, some of the datasets only present the aggregated annotation, and as result, we are unable to compute some of the ICR scores for these datasets. Table \ref{tab:gwet-datasets} show us the comparison between our datasets and some other previous works, with additional information on the number of annotated texts and the number of toxicity label categories. 

\begin{figure}[h]
\[ n = \frac{\frac{z^2p(1-p)}{e^2}}{1+\left(\frac{z^2p(1-p)}{e^2N}\right)} \]

\caption{This equation is used to calculate sample size $n$, where $z$ represents the Z-score associated with the confidence level, $p$ is the probability of a positive label, $e$ is the margin of error, and $N$ is the population size.}
\label{fig:sample-equation}
\end{figure}

While the number of texts in our datasets may seem relatively low compared to others, Equation in the Figure \ref{fig:sample-equation} shows that with a population of 42,846 texts, under the assumption that 20\% of the scraped texts were toxic, and setting the 95\% confidence level ($\alpha=0.05$) with a 5\% margin error, we find that the minimum number of required samples to represent the population is 245 texts. This showcase that while relatively small, our sample size is statistically representative.

\begin{table}[h]
\centering
\resizebox{0.65\columnwidth}{!}{
\begin{tabular}[ht]{llcc}
\toprule
\textbf{Dataset} & \textbf{details}& \textbf{Gwet's AC1}& \textbf{Fleiss Kappa} \\
\midrule
\citet{waseem-hovy-2016-hateful} & \begin{tabular}[c]{@{}l@{}}\textbullet\ \#texts: 6,654\\  \textbullet\ categories: 2 \end{tabular} & 0.78 & 0.57 \\

\hline
\textbf{Ours} & \begin{tabular}[c]{@{}l@{}}\textbullet\ \#texts: 250\\ \textbullet\ categories: 2 \end{tabular} & 0.61 & - \\

\hline
\citet{Davidson_Warmsley_Macy_Weber_2017} & \begin{tabular}[c]{@{}l@{}}\textbullet\ \#texts: 22,807\\ \textbullet\ categories: 3 \end{tabular} & - & 0.55 \\

\hline
\citet{haber-etal-2023-improving} & \begin{tabular}[c]{@{}l@{}}\textbullet\ \#texts: 15,000\\ \textbullet\ categories: 2 \end{tabular} & - & 0.31 \\

\hline
\citet{kumar-2021} & \begin{tabular}[c]{@{}l@{}}\textbullet\ \#texts: 107,620\\ \textbullet\ categories: 2 \end{tabular} & 0.27 & 0.26 \\

\bottomrule
\end{tabular}
}
\caption{The distribution of text that annotated by one or more annotators.}
\label{tab:gwet-datasets}
\end{table}

\newpage
\section{Dataset Properties}
\subsection{Annotation Statistics}
Table \ref{tab:annot-per-text} shows more fine-grained distribution on number of texts annotated by number of annotators.
\label{app:annotation-statistics}
\begin{table}[h]
\small
\centering
\resizebox{0.33\columnwidth}{!}{
\begin{tabular}[ht]{ccc}
\toprule
\textbf{\#annotators} & \textbf{\#texts}& \textbf{\% of total}\\
\midrule
1&15,748&55.36\\
2&7,907&27.79\\
3&2,352&8.27\\
4&1,755&6.17\\
5&21&0.07\\
6&215&0.76\\
7&1&0.0\\
11&26&0.09\\
12&2&0.01\\
13&150&0.53\\
14&1&0.0\\
15&146&0.51\\
16&2&0.01\\
17&97&0.34\\
19&25&0.09\\
\bottomrule
\end{tabular}
}
\caption{The distribution of text that annotated by one or more annotators.}
\label{tab:annot-per-text}

\end{table}

\subsection{Label Statistics}
\label{app:label-statistics}

Table \ref{tab:label-stat0} shows more detailed toxicity and polarization label distribution under different aggregation setup, while Table \ref{tab:label-stat} and Table \ref{tab:label-stat2} respectively shows the statistics of labeled data for toxicity types and related to election. \textbf{Any} aggregation is where an entry is labeled as positive if at least one annotator flags it, and \textbf{Consensus} aggregation is where we only consider texts with 100\% agreement of annotation.

\begin{table}[h]
\centering
\small
\begin{tabular}{@{\extracolsep{5pt}}clcccccc}
\hline
\multirow{2}{*}{\textbf{\#coder(s)}} & \multirow{2}{*}{\textbf{aggregation setup}} & \multicolumn{3}{c}{\textbf{Toxicity}} & \multicolumn{3}{c}{\textbf{Polarization}}\\ \cline{3-5} \cline{6-8}
 & \multicolumn{1}{c}{} & \#toxic & \multicolumn{1}{l}{\#non-toxic} & \multicolumn{1}{l}{Total} & \multicolumn{1}{l}{\#polarizing} & \multicolumn{1}{l}{\#non-polarizing} & \multicolumn{1}{l}{Total} \\ \hline
1 & - & 689 & 15,059 & 15,748 & 2,679 & 13,069 & 15,748 \\ \hline
\multirow{3}{*}{2+} & Majority & 1,467 & 9,394 & 10,861 & 1,132 & 8,847 & 9,969 \\
 & Any & 4,684 & 8,116 & 12,700 & 5,286 & 7,414 & 12,700\\
 & Consensus & 726 & 8,116 & 8,842 & 664 & 7,414 & 8,078\\ \hline
\end{tabular}
\caption{Number of toxic and polarizing texts based on several aggregation setup.}
\label{tab:label-stat0}
\vspace{-10pt}
\end{table}

\begin{table}[h]
\centering
\small
\begin{tabular}{@{\extracolsep{5pt}}clccccc}
\hline
\multirow{2}{*}{\textbf{\#coder(s)}} & \multirow{2}{*}{\textbf{aggregation setup}} & \multicolumn{5}{c}{\textbf{Toxicity Types}} \\ \cline{3-7}
 & \multicolumn{1}{c}{} & \#insults & \multicolumn{1}{l}{\#threat} & \multicolumn{1}{l}{\#profanity} & \multicolumn{1}{l}{\#identity-attack} & \multicolumn{1}{l}{\#sexually-explicit} \\ \hline
1 & - & 326 & 63 & 105 & 318 & 6 \\ \hline
\multirow{3}{*}{2+} & Majority & 422 & 25 & 155 & 455 & 44 \\
 & Any & 2,593 & 1,029 & 1,158 & 2,201 & 241\\
 & Consensus & 188 & 9 & 57 & 183 & 8\\ \hline
\end{tabular}
\caption{Number of texts per toxic types based on several aggregation setup. Keep in mind that one texts can contain multiple toxicity types.}
\label{tab:label-stat}
\vspace{-10pt}
\end{table}

\begin{table}[h]
\centering
\small
\begin{tabular}{@{\extracolsep{5pt}}clccc}
\hline
\multirow{2}{*}{\textbf{\#coder(s)}} & \multirow{2}{*}{\textbf{aggregation setup}} & \multicolumn{3}{c}{\textbf{Related to Election}} \\ \cline{3-5} 
 & \multicolumn{1}{c}{} & \#related & \multicolumn{1}{l}{\#not-related} & \multicolumn{1}{l}{Total}\\ \hline
1 & - & 922 & 14,826 & 15,748  \\ \hline
\multirow{3}{*}{2+} & Majority & 1,010 & 10,761 & 11,771  \\
 & Any & 2,403 & 10,297 & 12,700 \\
 & Consensus & 719 & 10,297 & 11,016\\ \hline
\end{tabular}
\caption{Number of texts with "Related to Election" label based on several aggregation setups.}
\label{tab:label-stat2}
\end{table}

\newpage
\section{Full Model Performance}
\label{app:full-model}

\subsection{Baseline Experiment}
\label{app:full-baseline}

\begin{table*}[ht]
\centering
\tiny
\setlength{\tabcolsep}{4pt} 
\renewcommand{\arraystretch}{1.2} 
\resizebox{\columnwidth}{!}{
\begin{tabular}{l | c c c c c c c c}
\hline
\textbf{Metric} & \textbf{IndoBERTweet} & \textbf{NusaBERT} & \textbf{Multi-e5} & \textbf{Llama3.1-8B} & \textbf{Aya23-8B} & \textbf{SeaLLMs-7B} & \textbf{GPT-4o} & \textbf{GPT-4o-mini} \\
\hline
\multicolumn{9}{c}{\textbf{Toxicity Detection}} \\
\hline
Accuracy                & $.844 \,\pm\, .008$ & $.841 \,\pm\, .005$ & $.834 \,\pm\, .007$ & $.646$ & $.750$ & $.512$ & $.829$ & $.819$ \\
Macro F1                & $.791 \,\pm\, .011$ & $.779 \,\pm\, .006$ & $.776 \,\pm\, .011$ & $.631$ & $.429$ & $.505$ & $.776$ & $.775$ \\
F1 (Class 0)            & $.896 \,\pm\, .006$ & $.896 \,\pm\, .004$ & $.890 \,\pm\, .005$ & $.705$ & $.857$ & $.565$ & $.885$ & $.875$ \\
F1 (Class 1)            & $.686 \,\pm\, .019$ & $.663 \,\pm\, .009$ & $.662 \,\pm\, .018$ & $.557$ & $.000$ & $.445$ & $.668$ & $.675$ \\
Precision (Class 1)     & $.692 \,\pm\, .022$ & $.704 \,\pm\, .018$ & $.675 \,\pm\, .015$ & $.405$ & $.000$ & $.311$ & $.649$ & $.613$ \\
Recall (Class 1)        & $.681 \,\pm\, .037$ & $.627 \,\pm\, .013$ & $.650 \,\pm\, .028$ & $.892$ & $.000$ & $.781$ & $.688$ & $.750$ \\
ROC AUC                 & $.790 \,\pm\, .015$ & $.769 \,\pm\, .006$ & $.773 \,\pm\, .013$ & --      & --      & --      & --      & --      \\
Precision-Recall AUC    & $.551 \,\pm\, .019$ & $.534 \,\pm\, .011$ & $.527 \,\pm\, .017$ & --      & --      & --      & --      & --      \\
\hline
\multicolumn{9}{c}{\textbf{Polarization Detection}} \\
\hline
Accuracy                & $.801 \,\pm\, .009$ & $.804 \,\pm\, .010$ & $.800 \,\pm\, .009$ & $.440$ & $.750$ & $.750$ & $.555$ & $.542$ \\
Macro F1                & $.731 \,\pm\, .013$ & $.732 \,\pm\, .016$ & $.735 \,\pm\, .011$ & $.440$ & $.429$ & $.411$ & $.553$ & $.540$ \\
F1 (Class 0)            & $.869 \,\pm\, .006$ & $.870 \,\pm\, .006$ & $.866 \,\pm\, .006$ & $.422$ & $.857$ & $.423$ & $.585$ & $.571$ \\
F1 (Class 1)            & $.593 \,\pm\, .020$ & $.593 \,\pm\, .026$ & $.604 \,\pm\, .018$ & $.457$ & $.000$ & $.399$ & $.521$ & $.508$ \\
Precision (Class 1)     & $.608 \,\pm\, .019$ & $.615 \,\pm\, .019$ & $.597 \,\pm\, .018$ & $.302$ & $.000$ & $.268$ & $.356$ & $.347$ \\
Recall (Class 1)        & $.579 \,\pm\, .027$ & $.574 \,\pm\, .038$ & $.612 \,\pm\, .025$ & $.942$ & $.000$ & $.781$ & $.968$ & $.946$ \\
ROC AUC                 & $.727 \,\pm\, .014$ & $.727 \,\pm\, .018$ & $.737 \,\pm\, .012$ & --      & --      & --      & --      & --      \\
Precision-Recall AUC    & $.457 \,\pm\, .017$ & $.460 \,\pm\, .022$ & $.462 \,\pm\, .016$ & --      & --      & --      & --      & --      \\
\hline
\end{tabular}}
\caption{Combined model performance on toxicity and polarization detection. ROC AUC and Precision-Recall AUC scores are not available for the LLMs.}
\label{tab:combined-metrics}
\end{table*}

\newpage
\subsection{Featural Experiment}
\label{app:full-featural}
\begin{table*}[ht]
\centering
\tiny
\setlength{\tabcolsep}{4pt} 
\renewcommand{\arraystretch}{1.2} 
\resizebox{\columnwidth}{!}{
\begin{tabular}{l | c c c c c}
\hline
\textbf{Metric} & \textbf{Baseline} & \textbf{Baseline (ANY)} & \textbf{Single Coder} & \textbf{Multiple Coders} & \textbf{Multiple Coders (ANY)} \\
\hline
\multicolumn{6}{c}{\textbf{Toxicity Detection}} \\
\hline
Accuracy                & $.844 \,\pm\, .008$ & $.769 \,\pm\, .012$ & $.831 \,\pm\, .006$ & $.827 \,\pm\, .014$ & $.828 \,\pm\, .010$ \\
Macro F1                & $.791 \,\pm\, .011$ & $.715 \,\pm\, .011$ & $.746 \,\pm\, .016$ & $.785 \,\pm\, .014$ & $.786 \,\pm\, .009$ \\
F1 (Class 0)            & $.896 \,\pm\, .006$ & $.839 \,\pm\, .011$ & $.893 \,\pm\, .003$ & $.880 \,\pm\, .012$ & $.880 \,\pm\, .008$ \\
F1 (Class 1)            & $.686 \,\pm\, .019$ & $.591 \,\pm\, .017$ & $.599 \,\pm\, .028$ & $.690 \,\pm\, .019$ & $.692 \,\pm\, .012$ \\
Precision (Class 1)     & $.692 \,\pm\, .022$ & $.532 \,\pm\, .023$ & $.736 \,\pm\, .011$ & $.628 \,\pm\, .033$ & $.627 \,\pm\, .024$ \\
Recall (Class 1)        & $.681 \,\pm\, .037$ & $.668 \,\pm\, .042$ & $.507 \,\pm\, .041$ & $.767 \,\pm\, .034$ & $.773 \,\pm\, .036$ \\
ROC AUC                 & $.790 \,\pm\, .015$ & $.735 \,\pm\, .014$ & $.723 \,\pm\, .018$ & $.807 \,\pm\, .013$ & $.810 \,\pm\, .011$ \\
Precision-Recall AUC    & $.551 \,\pm\, .019$ & $.438 \,\pm\, .015$ & $.496 \,\pm\, .019$ & $.539 \,\pm\, .023$ & $.541 \,\pm\, .014$ \\
\hline
\multicolumn{6}{c}{\textbf{Polarization Detection}} \\
\hline
Accuracy                & $.801 \,\pm\, .009$ & $.792 \,\pm\, .006$ & $.796 \,\pm\, .006$ & $.787 \,\pm\, .005$ & $.767 \,\pm\, .004$ \\
Macro F1                & $.731 \,\pm\, .013$ & $.736 \,\pm\, .006$ & $.736 \,\pm\, .008$ & $.674 \,\pm\, .011$ & $.706 \,\pm\, .007$ \\
F1 (Class 0)            & $.869 \,\pm\, .006$ & $.857 \,\pm\, .006$ & $.862 \,\pm\, .004$ & $.866 \,\pm\, .003$ & $.840 \,\pm\, .004$ \\
F1 (Class 1)            & $.593 \,\pm\, .020$ & $.614 \,\pm\, .012$ & $.610 \,\pm\, .012$ & $.481 \,\pm\, .021$ & $.572 \,\pm\, .016$ \\
Precision (Class 1)     & $.608 \,\pm\, .019$ & $.572 \,\pm\, .013$ & $.585 \,\pm\, .012$ & $.617 \,\pm\, .019$ & $.528 \,\pm\, .008$ \\
Recall (Class 1)        & $.579 \,\pm\, .027$ & $.664 \,\pm\, .037$ & $.637 \,\pm\, .019$ & $.395 \,\pm\, .030$ & $.625 \,\pm\, .043$ \\
ROC AUC                 & $.727 \,\pm\, .014$ & $.749 \,\pm\, .011$ & $.743 \,\pm\, .009$ & $.657 \,\pm\, .012$ & $.719 \,\pm\, .014$ \\
Precision-Recall AUC    & $.457 \,\pm\, .017$ & $.464 \,\pm\, .009$ & $.463 \,\pm\, .011$ & $.395 \,\pm\, .012$ & $.424 \,\pm\, .011$ \\
\hline
\end{tabular}}
\caption{Performance of IndoBERTweet variants on toxicity and polarization detection.}
\label{tab:indoBERTweet-variants}
\end{table*}

\subsection{Demographical}
\label{app:full-demo}
\subsubsection{IndoBERTweet}
\begin{table*}[h]
\centering
\small
\renewcommand{\arraystretch}{1.2} 
\begin{adjustbox}{max width=\textwidth}
\begin{tabular}{l | c c c c c c c c}
\hline
\textbf{Model} & \textbf{Accuracy} & \textbf{Macro F1} & \textbf{F1 (Class 0)} & \textbf{F1 (Class 1)} & \textbf{Precision (Class 1)} & \textbf{Recall (Class 1)} & \textbf{ROC AUC} & \textbf{PR AUC} \\
\hline
\multicolumn{9}{c}{\textbf{Toxicity Detection}} \\
\hline
Age Group                 & $.803 \pm .008$ & $.774 \pm .006$ & $.855 \pm .008$ & $.692 \pm .008$ & $.692 \pm .018$ & $.693 \pm .023$ & $.774 \pm .007$ & $.578 \pm .009$ \\
Baseline                  & $.680 \pm .007$ & $.405 \pm .002$ & $.809 \pm .005$ & $.000 \pm .000$ & $.000 \pm .000$ & $.000 \pm .000$ & $.500 \pm .000$ & $.320 \pm .007$ \\
Disability                & $.788 \pm .011$ & $.757 \pm .008$ & $.844 \pm .011$ & $.670 \pm .008$ & $.671 \pm .025$ & $.671 \pm .027$ & $.757 \pm .008$ & $.555 \pm .010$ \\
Domicile                  & $.808 \pm .007$ & $.773 \pm .008$ & $.862 \pm .006$ & $.684 \pm .015$ & $.724 \pm .020$ & $.650 \pm .040$ & $.766 \pm .013$ & $.582 \pm .005$ \\
Ethnicity                 & $.825 \pm .008$ & $.797 \pm .011$ & $.873 \pm .006$ & $.721 \pm .018$ & $.737 \pm .020$ & $.707 \pm .036$ & $.794 \pm .013$ & $.615 \pm .017$ \\
Ethnicity-Domicile-Religion & $.832 \pm .006$ & $.806 \pm .004$ & $.877 \pm .007$ & $.735 \pm .004$ & $.744 \pm .023$ & $.728 \pm .022$ & $.805 \pm .003$ & $.628 \pm .009$ \\
Gender                    & $.792 \pm .008$ & $.762 \pm .005$ & $.847 \pm .009$ & $.676 \pm .009$ & $.675 \pm .021$ & $.679 \pm .029$ & $.762 \pm .006$ & $.561 \pm .010$ \\
LGBT                      & $.788 \pm .010$ & $.756 \pm .008$ & $.844 \pm .010$ & $.667 \pm .011$ & $.672 \pm .021$ & $.664 \pm .032$ & $.755 \pm .009$ & $.553 \pm .009$ \\
Education                 & $.798 \pm .008$ & $.768 \pm .006$ & $.851 \pm .009$ & $.684 \pm .011$ & $.687 \pm .021$ & $.683 \pm .034$ & $.768 \pm .008$ & $.570 \pm .010$ \\
President Vote Leaning    & $.799 \pm .008$ & $.765 \pm .005$ & $.854 \pm .008$ & $.677 \pm .008$ & $.698 \pm .019$ & $.657 \pm .026$ & $.761 \pm .006$ & $.568 \pm .007$ \\
Religion                  & $.796 \pm .010$ & $.766 \pm .008$ & $.850 \pm .009$ & $.682 \pm .009$ & $.682 \pm .023$ & $.683 \pm .023$ & $.766 \pm .008$ & $.567 \pm .011$ \\
Employment Status         & $.793 \pm .010$ & $.764 \pm .006$ & $.847 \pm .011$ & $.681 \pm .005$ & $.674 \pm .026$ & $.689 \pm .025$ & $.765 \pm .004$ & $.563 \pm .011$ \\
\hline
\multicolumn{9}{c}{\textbf{Polarization Detection}} \\
\hline
Age Group                 & $.846 \pm .005$ & $.709 \pm .004$ & $.908 \pm .004$ & $.509 \pm .008$ & $.596 \pm .025$ & $.445 \pm .008$ & $.689 \pm .003$ & $.365 \pm .015$ \\
Baseline                  & $.820 \pm .010$ & $.450 \pm .003$ & $.901 \pm .006$ & $.000 \pm .000$ & $.000 \pm .000$ & $.000 \pm .000$ & $.500 \pm .000$ & $.180 \pm .010$ \\
Disability                & $.836 \pm .005$ & $.687 \pm .009$ & $.903 \pm .004$ & $.472 \pm .019$ & $.562 \pm .027$ & $.407 \pm .022$ & $.669 \pm .009$ & $.336 \pm .020$ \\
Domicile                  & $.850 \pm .005$ & $.716 \pm .003$ & $.911 \pm .004$ & $.522 \pm .008$ & $.612 \pm .035$ & $.457 \pm .019$ & $.696 \pm .005$ & $.377 \pm .016$ \\
Ethnicity                 & $.857 \pm .005$ & $.738 \pm .005$ & $.915 \pm .003$ & $.561 \pm .009$ & $.632 \pm .039$ & $.506 \pm .018$ & $.721 \pm .005$ & $.408 \pm .018$ \\
Ethnicity-Domicile-Religion & $.864 \pm .004$ & $.750 \pm .008$ & $.919 \pm .003$ & $.582 \pm .016$ & $.655 \pm .040$ & $.525 \pm .019$ & $.732 \pm .007$ & $.429 \pm .024$ \\
Gender                    & $.838 \pm .007$ & $.695 \pm .011$ & $.904 \pm .005$ & $.487 \pm .022$ & $.566 \pm .029$ & $.429 \pm .032$ & $.678 \pm .012$ & $.346 \pm .021$ \\
LGBT                      & $.837 \pm .006$ & $.684 \pm .007$ & $.904 \pm .004$ & $.465 \pm .015$ & $.569 \pm .028$ & $.393 \pm .011$ & $.664 \pm .006$ & $.333 \pm .019$ \\
Education                 & $.844 \pm .007$ & $.707 \pm .006$ & $.907 \pm .005$ & $.507 \pm .013$ & $.588 \pm .024$ & $.448 \pm .032$ & $.689 \pm .011$ & $.362 \pm .010$ \\
President Vote Leaning    & $.847 \pm .004$ & $.708 \pm .010$ & $.909 \pm .003$ & $.506 \pm .019$ & $.602 \pm .032$ & $.437 \pm .015$ & $.687 \pm .008$ & $.365 \pm .023$ \\
Religion                  & $.844 \pm .006$ & $.710 \pm .006$ & $.907 \pm .004$ & $.512 \pm .009$ & $.588 \pm .027$ & $.455 \pm .022$ & $.692 \pm .008$ & $.366 \pm .012$ \\
Employment Status         & $.836 \pm .009$ & $.689 \pm .012$ & $.902 \pm .006$ & $.476 \pm .022$ & $.559 \pm .009$ & $.416 \pm .036$ & $.672 \pm .015$ & $.338 \pm .013$ \\
\hline
\end{tabular}
\end{adjustbox}
\caption{Performance of IndoBERTweet demographic-aware models on toxicity and polarization detection.}
\label{tab:full-demo-bert}
\end{table*}

\newpage
\subsubsection{GPT-4o-mini}
\begin{table*}[h]
\centering
\small
\renewcommand{\arraystretch}{1.2} 
\begin{adjustbox}{max width=\textwidth}
\begin{tabular}{l | c c c c c c}
\hline
\textbf{Model} & \textbf{Accuracy} & \textbf{Macro F1} & \textbf{F1 (Class 0)} & \textbf{F1 (Class 1)} & \textbf{Precision (Class 1)} & \textbf{Recall (Class 1)} \\
\hline
\multicolumn{7}{c}{\textbf{Toxicity Detection}} \\
\hline
Age Group                 & $.804$ & $.788$ & $.846$ & $.730$ & $.710$ & $.752$ \\
Baseline                  & $.806$ & $.790$ & $.847$ & $.732$ & $.712$ & $.753$ \\
Disability                & $.804$ & $.789$ & $.846$ & $.731$ & $.710$ & $.754$ \\
Domicile                  & $.806$ & $.791$ & $.848$ & $.734$ & $.713$ & $.756$ \\
Ethnicity                 & $.805$ & $.789$ & $.847$ & $.731$ & $.711$ & $.753$ \\
Ethnicity-Domicile-Religion & $.807$ & $.797$ & $.841$ & $.753$ & $.687$ & $.834$ \\
Gender                    & $.804$ & $.789$ & $.846$ & $.731$ & $.710$ & $.754$ \\
LGBT                      & $.805$ & $.790$ & $.847$ & $.732$ & $.712$ & $.754$ \\
Education                 & $.805$ & $.790$ & $.847$ & $.732$ & $.712$ & $.753$ \\
President Vote Leaning    & $.805$ & $.790$ & $.847$ & $.732$ & $.712$ & $.754$ \\
Religion                  & $.804$ & $.789$ & $.846$ & $.731$ & $.711$ & $.752$ \\
Employment Status         & $.806$ & $.790$ & $.847$ & $.733$ & $.712$ & $.755$ \\
\hline
\multicolumn{7}{c}{\textbf{Polarization Detection}} \\
\hline
Age Group                 & $.527$ & $.527$ & $.545$ & $.509$ & $.346$ & $.967$ \\
Baseline                  & $.530$ & $.530$ & $.547$ & $.513$ & $.349$ & $.968$ \\
Disability                & $.529$ & $.528$ & $.546$ & $.510$ & $.346$ & $.967$ \\
Domicile                  & $.534$ & $.534$ & $.551$ & $.516$ & $.352$ & $.967$ \\
Ethnicity                 & $.535$ & $.534$ & $.552$ & $.517$ & $.352$ & $.968$ \\
Ethnicity-Domicile-Religion & $.542$ & $.540$ & $.565$ & $.516$ & $.352$ & $.962$ \\
Gender                    & $.529$ & $.528$ & $.546$ & $.510$ & $.346$ & $.967$ \\
LGBT                      & $.535$ & $.534$ & $.551$ & $.517$ & $.353$ & $.968$ \\
Education                 & $.531$ & $.531$ & $.548$ & $.514$ & $.350$ & $.968$ \\
President Vote Leaning    & $.528$ & $.527$ & $.545$ & $.509$ & $.346$ & $.966$ \\
Religion                  & $.534$ & $.534$ & $.551$ & $.517$ & $.353$ & $.968$ \\
Employment Status         & $.529$ & $.528$ & $.546$ & $.510$ & $.346$ & $.967$ \\
\hline
\end{tabular}
\end{adjustbox}
\caption{Performance of GPT-4o-mini demographic-aware models on toxicity and polarization detection.}
\label{tab:full-demo-gpt}
\end{table*}

\newpage
\subsection{Demographic + Featural}
\label{app:full-combined}
\begin{table*}[h]
\centering
\small
\renewcommand{\arraystretch}{1.2} 
\begin{adjustbox}{max width=\textwidth}
\resizebox{\columnwidth}{!}{
\begin{tabular}{l | c c c c c c c c}
\hline
\textbf{Model} & \textbf{Accuracy} & \textbf{Macro F1} & \textbf{F1 (Class 0)} & \textbf{F1 (Class 1)} & \textbf{Precision (Class 1)} & \textbf{Recall (Class 1)} & \textbf{ROC AUC} & \textbf{PR AUC} \\
\hline
\multicolumn{9}{c}{\textbf{Toxicity Detection}} \\
\hline
IndoBERTweet            & $.844 \pm .008$ & $.791 \pm .011$ & $.896 \pm .006$ & $.686 \pm .019$ & $.692 \pm .022$ & $.681 \pm .037$ & $.790 \pm .015$ & $.551 \pm .019$ \\
Best-featural           & $.872 \pm .008$ & $.828 \pm .011$ & $.915 \pm .005$ & $.741 \pm .018$ & $.749 \pm .019$ & $.735 \pm .033$ & $.826 \pm .015$ & $.617 \pm .020$ \\
Best-demo only          & $.832 \pm .006$ & $.806 \pm .004$ & $.877 \pm .007$ & $.735 \pm .004$ & $.744 \pm .023$ & $.728 \pm .022$ & $.805 \pm .003$ & $.628 \pm .009$ \\
\hline
Age Group              & $.818 \pm .005$ & $.790 \pm .003$ & $.867 \pm .006$ & $.714 \pm .006$ & $.720 \pm .023$ & $.710 \pm .024$ & $.790 \pm .004$ & $.604 \pm .010$ \\
Baseline                & $.680 \pm .007$ & $.405 \pm .002$ & $.809 \pm .005$ & $.000 \pm .000$ & $.000 \pm .000$ & $.000 \pm .000$ & $.500 \pm .000$ & $.320 \pm .007$ \\
Disability              & $.808 \pm .007$ & $.782 \pm .002$ & $.857 \pm .009$ & $.707 \pm .008$ & $.693 \pm .030$ & $.724 \pm .041$ & $.786 \pm .006$ & $.589 \pm .008$ \\
Domicile                & $.836 \pm .006$ & $.809 \pm .006$ & $.881 \pm .007$ & $.737 \pm .012$ & $.761 \pm .034$ & $.718 \pm .048$ & $.805 \pm .012$ & $.635 \pm .005$ \\
Ethnicity               & $.837 \pm .007$ & $.812 \pm .007$ & $.881 \pm .006$ & $.744 \pm .010$ & $.750 \pm .020$ & $.739 \pm .018$ & $.811 \pm .006$ & $.637 \pm .015$ \\
Ethnicity-Domicile-Religion & $.850 \pm .005$ & $.827 \pm .004$ & $.890 \pm .005$ & $.765 \pm .004$ & $.768 \pm .016$ & $.762 \pm .015$ & $.827 \pm .004$ & $.661 \pm .007$ \\
Gender                  & $.813 \pm .006$ & $.788 \pm .005$ & $.861 \pm .007$ & $.714 \pm .009$ & $.701 \pm .026$ & $.730 \pm .033$ & $.791 \pm .006$ & $.597 \pm .012$ \\
LGBT                    & $.811 \pm .010$ & $.784 \pm .008$ & $.861 \pm .009$ & $.708 \pm .008$ & $.703 \pm .022$ & $.713 \pm .019$ & $.785 \pm .008$ & $.593 \pm .011$ \\
Education     & $.814 \pm .008$ & $.788 \pm .006$ & $.861 \pm .009$ & $.716 \pm .004$ & $.701 \pm .027$ & $.733 \pm .024$ & $.792 \pm .003$ & $.599 \pm .012$ \\
President Vote Leaning  & $.824 \pm .006$ & $.797 \pm .006$ & $.872 \pm .006$ & $.722 \pm .009$ & $.733 \pm .021$ & $.713 \pm .022$ & $.795 \pm .006$ & $.614 \pm .012$ \\
Religion                & $.815 \pm .008$ & $.790 \pm .006$ & $.862 \pm .009$ & $.717 \pm .007$ & $.704 \pm .028$ & $.733 \pm .026$ & $.793 \pm .005$ & $.601 \pm .013$ \\
Employment Status        & $.811 \pm .008$ & $.786 \pm .007$ & $.859 \pm .009$ & $.713 \pm .012$ & $.694 \pm .024$ & $.735 \pm .042$ & $.791 \pm .011$ & $.594 \pm .010$ \\
\hline
\multicolumn{9}{c}{\textbf{Polarization Detection}} \\
\hline
IndoBERTweet            & $.801 \pm .009$ & $.731 \pm .013$ & $.869 \pm .006$ & $.593 \pm .020$ & $.608 \pm .019$ & $.579 \pm .027$ & $.727 \pm .014$ & $.457 \pm .017$ \\
Best-featural           & $.820 \pm .009$ & $.757 \pm .014$ & $.881 \pm .006$ & $.633 \pm .022$ & $.645 \pm .020$ & $.622 \pm .032$ & $.754 \pm .016$ & $.496 \pm .021$ \\
Best-demo only          & $.864 \pm .004$ & $.750 \pm .008$ & $.919 \pm .003$ & $.582 \pm .016$ & $.655 \pm .040$ & $.525 \pm .019$ & $.732 \pm .007$ & $.429 \pm .024$ \\
\hline
Age Group             & $.818 \pm .009$ & $.760 \pm .012$ & $.877 \pm .006$ & $.643 \pm .019$ & $.656 \pm .020$ & $.632 \pm .025$ & $.757 \pm .013$ & $.510 \pm .020$ \\
Baseline                & $.739 \pm .007$ & $.425 \pm .002$ & $.850 \pm .004$ & $.000 \pm .000$ & $.000 \pm .000$ & $.000 \pm .000$ & $.500 \pm .000$ & $.261 \pm .007$ \\
Disability              & $.804 \pm .009$ & $.744 \pm .016$ & $.868 \pm .006$ & $.619 \pm .027$ & $.627 \pm .019$ & $.612 \pm .038$ & $.742 \pm .019$ & $.485 \pm .025$ \\
Domicile                & $.849 \pm .008$ & $.801 \pm .011$ & $.898 \pm .006$ & $.704 \pm .017$ & $.719 \pm .014$ & $.690 \pm .026$ & $.797 \pm .012$ & $.577 \pm .018$ \\
Ethnicity               & $.849 \pm .009$ & $.804 \pm .010$ & $.898 \pm .007$ & $.710 \pm .013$ & $.711 \pm .018$ & $.710 \pm .020$ & $.804 \pm .010$ & $.580 \pm .015$ \\
Ethnicity-Domicile-Religion & $.871 \pm .006$ & $.830 \pm .008$ & $.913 \pm .004$ & $.748 \pm .013$ & $.759 \pm .012$ & $.738 \pm .021$ & $.827 \pm .010$ & $.628 \pm .016$ \\
Gender                  & $.804 \pm .010$ & $.741 \pm .014$ & $.869 \pm .007$ & $.614 \pm .024$ & $.632 \pm .017$ & $.599 \pm .044$ & $.738 \pm .018$ & $.483 \pm .020$ \\
LGBT                    & $.798 \pm .006$ & $.738 \pm .013$ & $.863 \pm .004$ & $.612 \pm .024$ & $.612 \pm .009$ & $.613 \pm .043$ & $.738 \pm .018$ & $.476 \pm .021$ \\
Education     & $.816 \pm .008$ & $.757 \pm .015$ & $.876 \pm .005$ & $.637 \pm .027$ & $.654 \pm .011$ & $.622 \pm .048$ & $.753 \pm .020$ & $.505 \pm .023$ \\
President Vote Leaning  & $.829 \pm .006$ & $.773 \pm .009$ & $.886 \pm .004$ & $.659 \pm .015$ & $.687 \pm .002$ & $.635 \pm .028$ & $.766 \pm .012$ & $.531 \pm .013$ \\
Religion                & $.829 \pm .009$ & $.771 \pm .013$ & $.886 \pm .006$ & $.655 \pm .021$ & $.692 \pm .018$ & $.623 \pm .035$ & $.762 \pm .015$ & $.529 \pm .019$ \\
Employment Status        & $.806 \pm .008$ & $.746 \pm .014$ & $.869 \pm .005$ & $.624 \pm .024$ & $.630 \pm .020$ & $.618 \pm .040$ & $.745 \pm .017$ & $.489 \pm .022$ \\
\hline
\end{tabular}}
\end{adjustbox}
\caption{Performance of IndoBERTweet-based models on toxicity and polarization detection.}
\label{tab:full-combined}
\end{table*}

\newpage
\section{LLMs' 2-Shot Setup Performance}
\label{app:llm-2shot}
\begin{table*}[ht]
\centering
\small
\setlength{\tabcolsep}{4pt} 
\renewcommand{\arraystretch}{1.2} 
\begin{tabular}{l | cc | cc | cc}
\hline
\multicolumn{7}{c}{\textbf{Toxicity Detection Performance}} \\
\hline
\multirow{2}{*}{\textbf{Model}} 
    & \multicolumn{2}{c|}{\textbf{Macro F1}} 
    & \multicolumn{2}{c|}{\textbf{Toxic F1}} 
    & \multicolumn{2}{c}{\textbf{Non-Toxic F1}} \\
  & 0‑shot & 2‑shot & 0‑shot & 2‑shot & 0‑shot & 2‑shot \\
\hline
GPT-4o-mini & \textbf{0.674} & 0.651 & \textbf{0.456} & 0.439 & \textbf{0.891} & 0.863 \\
Llama3.1-8B   & \textbf{0.511} & 0.483 & \textbf{0.280} & 0.262 & \textbf{0.742} & 0.704 \\
SeaLLMs-7B    & 0.384 & \textbf{0.454} & 0.185 & \textbf{0.236} & 0.583 & \textbf{0.673} \\
Aya23-8B      & 0.536 & \textbf{0.607} & 0.114 & \textbf{0.336} & 0.958 & \textbf{0.878} \\
\hline
\end{tabular}
\caption{Toxicity detection performance of LLMs in 0‑shot and 2‑shot setups. \textbf{Bolded} values highlight the better performing setup (0-shot vs 2-shot) based on the specific metric.}
\label{tab:tox-performance-gap-sidebyside}
\end{table*}

\begin{table*}[ht]
\centering
\small
\setlength{\tabcolsep}{4pt} 
\renewcommand{\arraystretch}{1.2} 
\begin{tabular}{l | cc | cc | cc}
\hline
\multicolumn{7}{c}{\textbf{Polarization Detection Performance}} \\
\hline
\multirow{2}{*}{\textbf{Model}} 
    & \multicolumn{2}{c|}{\textbf{Macro F1}} 
    & \multicolumn{2}{c|}{\textbf{Polar F1}} 
    & \multicolumn{2}{c}{\textbf{Non-Polar F1}} \\
  & 0‑shot & 2‑shot & 0‑shot & 2‑shot & 0‑shot & 2‑shot \\
\hline
GPT-4o-mini & 0.536 & \textbf{0.609} & 0.450 & \textbf{0.512} & 0.621 & \textbf{0.706} \\
Llama3.1-8B   & 0.370 & \textbf{0.485} & 0.306 & \textbf{0.357} & 0.434 & \textbf{0.613} \\
SeaLLMs-7B    & 0.354 & \textbf{0.455} & 0.441 & \textbf{0.343} & 0.267 & \textbf{0.566} \\
Aya23-8B      & 0.466 & \textbf{0.526} & 0.013 & \textbf{0.310} & \textbf{0.919} & 0.743 \\
\hline
\end{tabular}
\caption{Polarization detection performance of LLMs in 0‑shot and 2‑shot setups.}
\label{tab:pol-performance-gap-sidebyside}
\end{table*}

Using a much smaller data subset (see Table \ref{tab:aggregate-labelcount}'s \textbf{2+} data count), we conducted a preliminary research. We show that for two of the highest performing LLMs (GPT-4o-mini and Llama3.1-8B), their performance degrades for toxicity detection (Table \ref{tab:tox-performance-gap-sidebyside}). Meanwhile, for polarization detection, their performance improves (Table \ref{tab:pol-performance-gap-sidebyside}). Due to this difference in behavior, we chose to prioritize the 0-shot setup instead.

\newpage
\section{IndoBERTweet Input Setup and GPT-4o-mini Prompts List}
\label{app:input-prompts}

Differing experiments require differing setup of the model's input. For IndoBERTweet, we leverage BERT's pre-training schematic and utilize the [SEP] token, following \citet{kumar-2021}'s setup. For GPT-4o-mini, we augment its input by pre-pending specific texts depending on the experiment. These augmentations are available at Table \ref{tab:prompt_templates}.

\begin{table*}[htbp]
  \centering
  \small
  \resizebox{\columnwidth}{!}{
  \begin{tabularx}{\textwidth}{p{3cm} X X}
    \toprule
    \textbf{Experiment} & \textbf{IndoBERTweet} & \textbf{GPT-4o-mini} \\
    \midrule
    Baseline &
    \{TEXT\} &
    "Answer only with [``toxic''/``polarizing''] or [``not toxic''/``not polarizing''].\newline
    Is this Indonesian text [toxic/polarizing]?\newline
    ``````\newline
    \{TEXT\}\newline
    `````` \\
    \midrule
    Featural &
    Nilai rata-rata [toksisitas/polarisasi]: \{VALUE\} [SEP] \{TEXT\} &
    "Answer only with [``toxic''/``polarizing''] or [``not toxic''/``not polarizing''].\newline
    Is this Indonesian text with a [toxicity/polarization] index (range of 0 to 1) of \{VALUE\} [toxic/polarizing]?\newline
    ``````\newline
    \{TEXT\}\newline
    `````` \\
    \midrule
    Demographical &
    "Informasi Demografis:\newline
    \{DEMOGRAPHIC\_CLASS\_1\}: \{DEMOGRAPHIC\_VALUE\_1\}\newline
    ...\newline
    \{DEMOGRAPHIC\_CLASS\_n\}: \{DEMOGRAPHIC\_VALUE\_n\} [SEP] \{TEXT\} &
    Answer only with [``toxic''/``polarizing''] or [``not toxic''/``not polarizing''].\newline
    You are an Indonesian citizen with the following demographic information:\newline
    \{DEMOGRAPHIC\_CLASS\_1\}: \{DEMOGRAPHIC\_VALUE\_1\}\newline
    ...\newline
    \{DEMOGRAPHIC\_CLASS\_n\}: \{DEMOGRAPHIC\_VALUE\_n\}\newline
    Is this Indonesian text [toxic/polarizing]?\newline
    ``````\newline
    \{TEXT\}\newline
    `````` \\
    \midrule
    Demographical and Featural &
    Informasi Demografis:\newline
    \{DEMOGRAPHIC\_CLASS\_1\}: \{DEMOGRAPHIC\_VALUE\_1\}\newline
    ...\newline
    \{DEMOGRAPHIC\_CLASS\_n\}: \{DEMOGRAPHIC\_VALUE\_n\}\newline
    Nilai rata-rata [toksisitas/polarisasi]: \{VALUE\} [SEP] \{TEXT\} &
    "Answer only with [``toxic''/``polarizing''] or [``not toxic''/``not polarizing''].\newline
    You are an Indonesian citizen with the following demographic information:\newline
    \{DEMOGRAPHIC\_CLASS\_1\}: \{DEMOGRAPHIC\_VALUE\_1\}\newline
    ...\newline
    \{DEMOGRAPHIC\_CLASS\_n\}: \{DEMOGRAPHIC\_VALUE\_n\}\newline
    Is this Indonesian text with a [toxicity/polarization] index (range of 0 to 1) of \{VALUE\} [toxic/polarizing]?\newline
    ``````\newline
    \{TEXT\}\newline
    `````` \\
    \bottomrule
  \end{tabularx}}
  \caption{Prompt templates for IndoBERTweet and GPT-4o-mini experiments.}
  \label{tab:prompt_templates}
\end{table*}

\newpage
\section{Predictor Model Performance}
\label{app:predictor-perf}
Performance of the predictor model on Section \ref{abl:featural} visible on Table \ref{tab:app-predictor-perf}. \textbf{AGG} represents the independent variable as a value between [0, 1]; while \textbf{ANY} represents the independent variable as a binary value of 0 or 1. Because of this, the predictor differs per setup, where on \textbf{(AGG)} the predictor is a regressor while on \textbf{ANY} it is a classifier. 

\begin{table}[ht]
\centering
\small
\setlength{\tabcolsep}{6pt} 
\renewcommand{\arraystretch}{1.1} 

\begin{tabular}{l | c c | l | c c}
\hline
\multicolumn{3}{c|}{\textbf{Toxicity}} & \multicolumn{3}{c}{\textbf{Polarization}} \\
\hline
\textbf{Metric} & \textbf{(Agg) Pred} & \textbf{(Any) Pred} & \textbf{Metric} & \textbf{(Agg) Pred} & \textbf{(Any) Pred} \\
\hline
MSE & 0.109 & --- & MSE & 0.072 & --- \\
MAE & 0.222 & --- & MAE & 0.163 & --- \\
F1$_0$ & --- & 0.831 & F1$_0$ & --- & 0.907 \\
F1$_1$ & --- & 0.649 & F1$_1$ & --- & 0.504 \\
ROC AUC & --- & 0.736 & ROC AUC & --- & 0.691 \\
\hline
\end{tabular}

\caption{Comparison of (Agg) and (Any) Predictor models for Toxicity and Polarization tasks.}
\label{tab:app-predictor-perf}
\end{table}

\section{GPT-4o's Persona}
\label{app:gpt-persona}

Table \ref{tab:gpt-4o-persona-toxic} and \ref{tab:gpt-4o-persona-polarized} present the highest ICR group score from each demographic. To compute the toxicity ICR score for a demographic group, we calculated the weighted average of Gwet’s AC1 scores for every pairwise combination between GPT-4o and annotators within the respective group, using the volume of text in each pair as the weight.


\begin{table}[h]
\centering
\begin{tabular}{llc}
\hline
\textbf{demographic} & \multicolumn{1}{l}{\textbf{group}} & \textbf{Toxicity ICR (avg)} \\ \hline
Ethnicity & Non-indigenous & 0.751 \\
Domicile & Greater Jakarta & 0.746 \\
Religion & Non-Islam & 0.743 \\
Disability & Yes & 0.734 \\
Age Group & Gen X & 0.731 \\
President Vote Leaning & Candidate No. 2 & 0.724 \\
Education & Postgraduate Degree & 0.715 \\
Job Status & Unemployed & 0.707 \\
Gender & Female & 0.694 \\ \hline
\end{tabular}
\caption{GPT-4o's most highest ICR score for toxicity.}
\label{tab:gpt-4o-persona-toxic}
\end{table}

\begin{table}[h]
\centering
\begin{tabular}{llc}
\hline
\textbf{demographic} & \multicolumn{1}{l}{\textbf{group}} & \textbf{Polarized ICR (avg)} \\ \hline
Domicile & Javanese-Region & 0.566 \\
President Vote Leaning & Unknown & 0.408 \\
Age Group & Gen-X & 0.182 \\
Education & Postgraduate Degree & 0.108 \\
Disability & No & 0.066 \\
Ethnicity & Indigenous & 0.065 \\
Job Status & Students & 0.061 \\
Gender & Female & 0.059 \\
Religion & Islam & 0.059 \\ \hline
\end{tabular}
\caption{GPT-4o's most highest ICR score for toxicity.}
\label{tab:gpt-4o-persona-polarized}
\end{table}

\newpage
\section{In-group vs Out-group Agreement Gap}
\label{app:aggrement-gap}
\begin{table}[ht]
  \centering
  \small
  \resizebox{\columnwidth}{!}{
  \begin{tabular}{|c|c|c|c|c|c|c|c|}
    \hline
    \textbf{index} & \textbf{demographic} & \textbf{group} & \textbf{toxic\_gwet} & \textbf{toxic\_gwet\_diff} & \textbf{polarize\_gwet} & \textbf{polarize\_gwet\_diff} & \textbf{support} \\ \hline
    0 & disability & no & .40 & .37 & .32 & .46 & 26 \\ \hline
    1 & disability & yes & .77 & .37 & .78 & .46 & 3 \\ \hline
    2 & general\_domicile & Non-Java & .23 & .25 & .48 & .16 & 6 \\ \hline
    3 & general\_domicile & Greater Jakarta & .59 & .22 & .50 & .19 & 10 \\ \hline
    4 & general\_domicile & Java Region & .23 & .22 & .44 & .03 & 2 \\ \hline
    5 & age group & Gen X & .63 & .21 & .33 & .00 & 3 \\ \hline
    6 & ethnicity2 & Non-Indigenous & .60 & .20 & .37 & .05 & 4 \\ \hline
    7 & ethnicity2 & Indigenous & .40 & .20 & .32 & .05 & 25 \\ \hline
    8 & job status & Unemployed & .59 & .18 & .44 & .13 & 3 \\ \hline
    9 & president vote leaning & 1 & .59 & .16 & .43 & .12 & 9 \\ \hline
    10 & general\_domicile & Sumatera & .56 & .13 & .43 & .08 & 7 \\ \hline
    11 & general\_domicile & Bandung & .56 & .13 & .62 & .28 & 4 \\ \hline
    12 & religion2 & Non-Islam & .52 & .11 & .41 & .12 & 9 \\ \hline
    13 & religion2 & Islam & .41 & .11 & .29 & .12 & 20 \\ \hline
    14 & education & Postgraduate Degree & .51 & .07 & .44 & .10 & 7 \\ \hline
    15 & president vote leaning & unknown & .51 & .07 & .39 & .05 & 3 \\ \hline
    16 & president vote leaning & 2 & .50 & .07 & .39 & .06 & 9 \\ \hline
    17 & job status & Students & .41 & .06 & .29 & .13 & 8 \\ \hline
    18 & president vote leaning & 3 & .38 & .06 & .23 & .15 & 8 \\ \hline
    19 & gender & F & .44 & .04 & .25 & .17 & 16 \\ \hline
    20 & gender & M & .40 & .04 & .42 & .17 & 13 \\ \hline
    21 & job status & Employed & .44 & .03 & .39 & .09 & 18 \\ \hline
    22 & age group & Gen Z & .44 & .02 & .28 & .14 & 12 \\ \hline
    23 & age group & Millennials & .43 & .02 & .41 & .13 & 14 \\ \hline
    24 & education & Bachelor/Diploma & .43 & .01 & .41 & .11 & 14 \\ \hline
    25 & education & Highschool Degree & .45 & .01 & .29 & .11 & 8 \\ \hline
  \end{tabular}}
  \caption{Demographic Agreement Scores. \textbf{ethnicity2} and \textbf{religion2} denote higher-level groupings of demographic information (e.g., Christians and Buddhists are grouped as "Non-Islam").}
  \label{tab:demo_agreement}
\end{table}

\appendix

\newpage

\newpage
\newpage

\end{document}